\documentclass[a4paper,10pt]{article}
\usepackage{amsthm}
\usepackage{amsmath}
\usepackage{amssymb}
\usepackage{natbib}
\usepackage[colorlinks,citecolor=blue,urlcolor=blue,filecolor=blue,backref=page]{hyperref}
\usepackage{graphicx}
\usepackage{mathpazo}
\usepackage{microtype} 
\usepackage{booktabs} 
\usepackage{fullpage}
\usepackage{xcolor}
\usepackage{algpseudocode,algorithm,algorithmicx}
\usepackage{makecell}

\newcommand{\MMD}{\ensuremath{\textrm{MMD}}}
\newcommand{\unobs}{U}
\newcommand{\unobss}{u}
\newcommand{\data}{X^\obs}
\newcommand{\E}{\mathbb{E}}
\newcommand{\F}{\mathcal{F}}

\newcommand{\normal}{\mathcal{N}}

\newcommand{\gammarand}{\mathrm{Gamma}}
\newcommand{\betarand}{\mathrm{Beta}}
\renewcommand{\b}[1]{\mathbf{#1}}
\renewcommand{\d}{\textrm{d}}
\newcommand{\obs}{\ensuremath{\mathrm{obs}}}
\newcommand{\rep}{\ensuremath{\mathrm{rep}}}

\newcommand{\p}{\textrm{Pr}}
\newcommand{\disc}{D}
\newcommand{\1}{\mathbf{1}}
\newcommand{\pplug}{\ensuremath{p_\textrm{plug-in}}}
\newcommand{\pprior}{\ensuremath{p_\textrm{prior}}}
\newcommand{\ppost}{\ensuremath{p_\textrm{post}}}
\renewcommand{\F}{\mathcal{F}}

\newcommand{\bx}{\textbf{x}}
\newcommand{\bA}{\textbf{A}}
\newcommand{\bB}{\textbf{B}}
\newcommand{\bI}{\textbf{I}}

\newcommand{\reals}{\mathbb{R}}

\newcommand{\bZ}{\mathbf{Z}}
\newcommand{\bX}{\mathbf{X}}

\newcommand{\bz}{\mathbf{z}}

\newcommand{\bu}{\mathbf{u}}
\newcommand{\bU}{\mathbf{U}}

\newcommand{\doubleexp}{\mathcal{L}}
\newcommand{\Dirichlet}{\mathrm{Dir}}
\newcommand{\categorial}{\mathrm{Cat}}
\newcommand{\g}{\,|\,} % given
\newcommand{\bfx}{\mathbf{x}}
\newcommand{\bfz}{\mathbf{z}}
\newcommand{\bfmu}{\boldsymbol{\mu}}
\newcommand{\bftau}{\boldsymbol{\tau}}
\newcommand{\bfepsilon}{\boldsymbol{\epsilon}}
\newcommand{\bfpsi}{\boldsymbol{\psi}}
\newcommand{\bfb}{\mathbf{b}}
\newcommand{\bfs}{\mathbf{s}}
\newcommand{\bfc}{\mathbf{c}}
\newcommand{\bfy}{\mathbf{y}}
\newcommand{\bfm}{\mathbf{m}}
\newcommand{\bfK}{\mathbf{K}}
\newcommand{\bfX}{\mathbf{X}}
\newcommand{\bfU}{\mathbf{U}}
\newcommand{\bfu}{\mathbf{u}}
\newcommand{\bLambda}{\Lambda}
\newcommand{\bzero}{\textbf{0}}
\newcommand\inv[1]{#1\raisebox{1.15ex}{$\scriptscriptstyle-\!1$}}
\newcommand\invhalf[1]{#1\raisebox{1.15ex}{$\scriptscriptstyle-\!1/2$}}

\newtheorem{stat}{Statement}
\newcommand{\prop}{Statement}

\newcommand{\imgpath}{./}%{figures4}%{../figures/png/}
\newcommand{\cut}[1]{}
\newcommand{\eq}{\!=\!}
\newcommand{\bfTheta}{{\boldsymbol{\Theta}}}
\newcommand{\bftheta}{{\boldsymbol{\theta}}}
\newcommand{\bQ}{\mathbf{Q}}
\newcommand{\bR}{\mathbf{R}}
\newcommand{\bpi}{{\boldsymbol{\pi}}}

\allowbreak
\allowdisplaybreaks

\newcommand{\addr}{School of Informatics, University of Edinburgh, UK}
\title{Model Criticism in Latent Space}

\author{Sohan Seth\thanks{\addr; seth@inf.ed.ac.uk} 
\and
Iain Murray\thanks{\addr; i.murray@ed.ac.uk}\and
Christopher K. I. Williams\thanks{\addr; and the Alan Turing Institute, London, UK;
ckiw@inf.ed.ac.uk} \thanks{SS and CW gratefully acknowledge the UK Engineering
and Physical Sciences Research Council (EP/K03197X/1) for funding this work.
The work of CW is supported by EPSRC grant EP/N510129/1 to the
Alan Turing Institute. 
Code used in the paper is available at \url{https://github.com/sohanseth/mcls}.
}}
\date{}

\makeatletter
\newcommand{\owntag}[2][\relax]{% \owntag[short label]{tag}
  \ifx#1\relax\relax\def\owntag@name{#2}\else\def\owntag@name{#1}\fi% base label
  \refstepcounter{equation}\tag{\theequation, #2}%
  \expandafter\ltx@label\expandafter{eq:\owntag@name}%
  \edef\@currentlabel{\theequation, #2}\expandafter\ltx@label\expandafter{Eq:\owntag@name}%
  \def\@currentlabel{#2}\expandafter\ltx@label\expandafter{tag:\owntag@name}%
}
\makeatother

\begin{document}
\maketitle
\begin{abstract}
Model criticism is usually carried out by assessing if replicated data
generated under the fitted model looks similar to the observed data, see e.g.\
\citet*[p.\,165]{gelman-carlin-stern-rubin-04}. This paper presents a method for
latent variable models by pulling back the data into the space of latent
variables, and carrying out model criticism in that space. Making use of a
model's structure enables a more direct assessment of the assumptions made in
the prior and likelihood. We demonstrate the method with examples of model
criticism in latent space applied to factor analysis, linear
dynamical systems and Gaussian processes.
\end{abstract}

\section{Introduction}
Model criticism is the process of assessing the goodness of fit between
some data and a statistical model of that data%
\footnote{Following \citet[p423]{ohagan-03} we prefer the term \emph{model
criticism} over \emph{model validation} and \emph{model checking}, as if ``all
models are strictly wrong'' it is impossible to validate a model, and model
criticism has a more active tone of looking to discover problems, compared to
model checking, which may seem a more passive activity that does not expect to
uncover any problems.}.
While model criticism uses goodness-of-fit tests to judge aspects of the
model, its general objective is to identify deficiencies in the model that can
lead to \emph{model extension} to address these deficiencies. The extended
model(s) can again be subjected to criticism, and the process continues until a
\emph{satisfactory} model is found \citep{ohagan-03}.
Model criticism is contrasted with \emph{model comparison} in that model
criticism assesses a single model, while model comparison deals with at least
two models to decide which model is a better fit. Model comparison can be
applied to compare the original and the extended model after model criticism
and extension \cite[p.\,2]{ohagan-03}.

Bayesian modelling has become an indispensable tool in statistical
learning, and it is being widely used to model complex signals, e.g.,
by \citet{Ratmann10576}. With its growing popularity, there is need
for model criticism in this framework.
Most work on model criticism makes use of the idea that ``if the model fits,
then replicated data generated under the model should look similar to observed
data'' \citep[p.\,165]{gelman-carlin-stern-rubin-04}.  In contrast, in this
paper we focus on a less well explored idea that for latent variable
models, we can probabilistically pull back the data into the space of the
latent variables, and carry out model criticism in that space.  We can
summarize this principle as that \emph{if the model fits, then posterior
inferences should match the prior assumptions}.

\begin{figure*}[t]
\centering
\includegraphics[width=\textwidth]{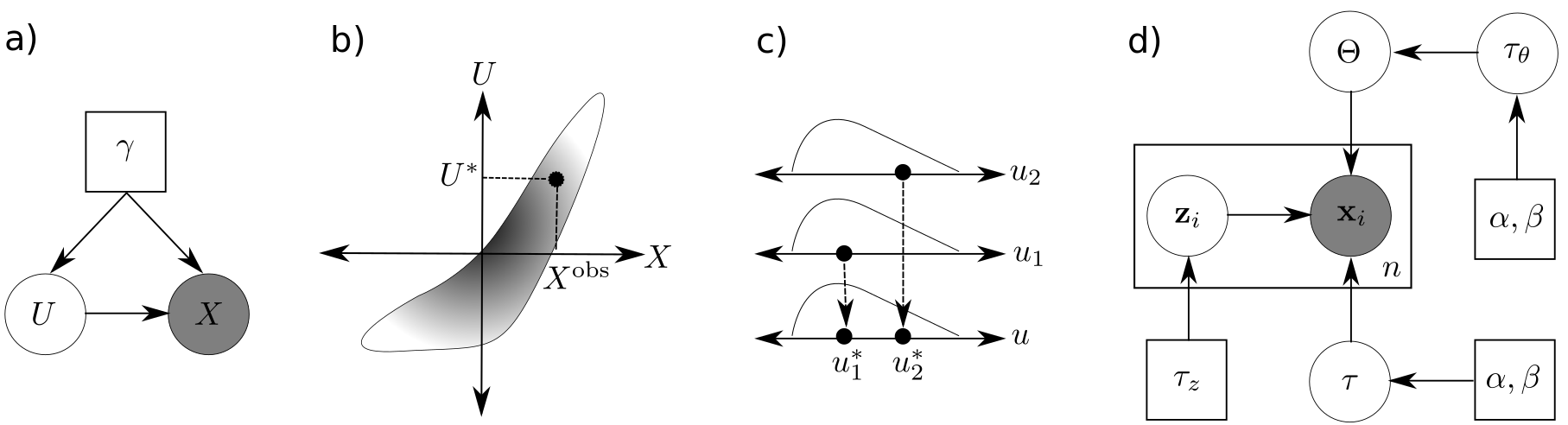}
\vspace*{-0.2in}
\caption{\textbf{a)}
A probabilistic model with observed variables $X$, unobserved variables
$\unobs$, and known parameters $\gamma$,
\textbf{b)}
Given the observed data $x^\obs$ from $P(X\g \gamma)$ and a posterior
sample $\unobss^\ast$ from $P(\unobs\g x^\obs,\gamma)$, $(x^\obs,
\unobss^\ast)$ is a joint sample from $P(X,\unobs\g \gamma)$, and therefore,
$\unobss^\ast$ is a sample from $P(\unobs\g \gamma)$ and $x^\obs$ is a
sample from $P(X\g \unobss^\ast,\gamma)$.
\textbf{c)}
If the prior (or part of it) factorizes into identical distributions, e.g.,
$P(\unobs \g  \gamma) \!=\! \prod_{k=1}^2 P_u(U_k\g \gamma)$, then posterior
sample $\{u_1^\ast,u_2^\ast\}$ is independent and identical sample from
$P_u(\cdot \g \gamma)$.  
\textbf{d)} A factor analysis model showing observed variables $X \!=\!
\{\bfx_i\}_{i=1}^n$, unobserved variables $\unobs \!=\!
\{\{\bfz_i\}_{i=1}^n,\Theta,\tau,\tau_\theta\}$, and known parameters
$\gamma\!=\!\{\alpha,\beta,\tau_z\}$. We test if
$\{z_{11}^\ast,z_{12}^\ast,\ldots\}$ is a sample from $P(Z\g \tau_z)$, and
$\{\theta_{11}^\ast, \theta_{12}^\ast, \ldots\}$ is a sample from
$P(\theta\g \tau_\theta^\ast)$.}
\label{fig:mc}
\end{figure*}

To elaborate, consider a model with observed variables $X$ and
unobserved variables $\unobs$ with joint distribution $P(X,\unobs\g
\gamma)$ where $\gamma$ are known parameters. In general $U$ may
contain latent variables $Z$, parameters $\Theta$, and hyperparameters
$\lambda$\footnote{For example, in the context of the Bayesian matrix
  factorization \citep{salakhutdinov2008b}, $X$ is the observed data
  matrix, $U=\{Z,\Theta,\lambda\}$ is the matrix of latent factors
  $Z$, the loading matrix $\Theta$, precision hyperparameters
  $\lambda$, and $\gamma$ denotes the parameters of the hyperpriors.}.
Given a sample $x^\obs$ from the marginal distribution $P(X\g \gamma)$, 
and a
single posterior sample $\unobss^\ast$ from the conditional
distribution $P(\unobs \g
x^\obs,\gamma)$, the joint sample $(x^\obs,\unobss^\ast)$ is a draw
from the distribution $P(X,\unobs\g \gamma)$.  This property can be
used to check the fit of the model in the latent space by checking if
$\unobss^\ast$ is a sample from the marginal distribution 
$P(\unobs\g \gamma)$. Testing a single sample
against a distribution, however, is not an effective approach.  But,
in many widely-used models, groups of unknown variables are
independently and identically distributed under the prior.  These
related variables are easily \emph{aggregated} together, giving a
simple test of the prior assumptions.  Figure~\ref{fig:mc} summarizes
the overall approach, which is justified in \S\ref{sec:latcritic}.

In comparison to model criticism in the observation space, comparing
$u^\ast$ with prior $P(U\g\gamma)$, provides an additional tool for model
criticism which does not require crafting an appropriate discrepancy measure,
generating replicate observations, and approximating the null distribution.
This approach also does not suffer from the ``double use'' of data
(see discussion in \S\ref{sec:obscritic}).
These points have also been made by \citet{YuanJohnson12}, but were applied to a
relatively small scale hierarchical linear model.  We develop the use of model
criticism in latent space for large scale and complex models, yielding new
insights and developments.  Specifically, we apply this approach to the
criticism of linear dynamical systems, factor analysis and Gaussian processes,
and discuss its connection to the observation space based approach.

The structure of the rest of the paper is as follows: in \S\ref{sec:obscritic}
we describe the methods of model criticism in observation space.
\S\ref{sec:latcritic} provides details of the argument for model criticism in
latent space and describes related work, and \S\ref{sec:results} shows results
from applying the method to the three examples. Table~\ref{table:notation}
describes the notations followed in the paper, and Table~\ref{table:dist} shows
the distributions used in the paper.

\begin{table}[h]
\centering
\scriptsize
\begin{tabular}{lll}
\toprule
Style & Explanation & Example \\
\midrule
Upper case italics & 
Random variable or a group of random variables &
$X, Z, U = \{U_1,\ldots,U_K\}$ \\
Lower case italics &
Realization of a random variable &
$\{x_i\}_{i=1}^n, u^\ast$ \\
Lower case bold &
Vectors, realization or random variable &
$\{\bx_i\}_{i=1}^n, P(\bz), \bu = (u_1,\ldots,u_K)^\top$ \\
Upper case bold &
Matrices, realization or random variable &
$\{\bX_i\}_{i=1}^n, P(\bZ), \bU = [\bu_1,\ldots,\bu_K]$ \\
$P(X)$ &
Distribution of random variable $X$ &
$X \sim P(X)$\\
$P(X\g y)$ &
Conditional distribution of random variable $X$ given $Y=y$ &
$X \sim P(X\g y)$\\
$p(x)$ &
Probability density function of r. v. $X$, abbreviation for $p_X(x)$ &
$p(y) = \int p(x,y) \d x$\\
$p(x \g y)$ &
Conditional density function of r. v. $X$ given $Y=y$, 
abbreviation for $p_{X\g Y}(x\g y)$ &
$p(y) = \int p(x\g y)p(y) \d x$\\
$\cdot \sim \cdot$ &
Distributed as &
$X^\rep \sim P(X), X \sim \normal(0,1)$\\
$\cdot^\ast$ &
A posterior sample &
$u^\ast, \bz^\ast$\\
$\cdot^\obs$ &
Observed data &
$x^\obs, X^\obs \sim P(X)$\\
$\cdot^\rep$ &
Replicate data &
$x^\rep, X^\rep \sim P(X)$\\
$\bzero$ &
Zero vector &
$\bx \sim \normal(\bzero, \bI)$\\
$\bI$ &
Identity matrix &
$\bx \sim \normal(\bzero, \bI)$\\
\bottomrule
\end{tabular}
\caption{Description of notation used in the paper.}
\label{table:notation}
\end{table}

\begin{table}[h]
\centering
\footnotesize
\begin{tabular}{lll}
\toprule
Name & Notation & Density \\
\midrule
Normal & $\normal(x\g \mu,\inv{\tau})$ & $\frac{{\tau}^{1/2}}{(2\pi)^{1/2}}\exp\left(-(\tau(x - \mu)^2)/2\right)$ \\[0.1in]
Multivariate normal & $\normal(\bfx\g \bfmu,\inv{\bftau})$  & $\frac{{\mathrm{det}(\bftau)}^{1/2}}{(2\pi)^{d/2}} \exp\left(-((\bfx-\bfmu)^\top\bftau(\bfx - \bfmu))/2\right)$ \\[0.1in]
Double exponential & $\doubleexp(x\g \mu,\tau)$ & $\frac{\tau}{2}\exp(-\tau|x-\mu|)$ \\[0.1in]
Gamma & $\gammarand(x\g \alpha,\beta)$ & $\frac{\beta^\alpha}{\Gamma(\alpha)}x^{\alpha-1}\exp(-\beta x)$ \\[0.1in]
Dirichlet & $\Dirichlet(\bfx\g \alpha)$ & $\frac{1}{B(\alpha)}{ \prod_{i=1}^K} x_i^{\alpha_i -1}$ \mbox{where} $x_i \in (0,1)$ \mbox{ and } $\sum_{i=1}^K x_i = 1$. \\
\bottomrule
\end{tabular}
\caption{List of distributions used in the paper.}
\label{table:dist}
\end{table}

\section{Model Criticism in Observation Space \label{sec:obscritic}}

A general approach of model criticism is to evaluate if replicated data
generated under the (fitted) model looks similar to observed data.  Consider
that we are modelling observed data $x^\obs$ with a latent variable model
parameterized by $\unobs$, i.e., we have defined the likelihood 
$p(x \g \unobss)$
and (optionally) a prior distribution $P(\unobs)$ over potential parameter
values.  The principle of model criticism in the observation space is to assess
if $x^\obs$ is a reasonable observation under the proposed model.
For example, given the \emph{maximum likelihood estimator} (or another point
estimate) $\hat{\unobss}$ of the parameters, one standard approach is to find
the \emph{plug-in} p-value \citep{Bayarri00}
\begin{align}
\pplug = \p(\disc(X^\rep, \hat{\unobss}) > \disc(x^\obs, \hat{\unobss})).
\end{align}
Here $\disc$ is called a \emph{discrepancy function} and it resembles a test
statistic in hypothesis testing, i.e., a larger value rejects the null
hypothesis or indicates incompatibility of data and model, and
$X^\rep$ is a
\emph{replicate observation} generated under the fitted model, i.e., 
$X^\rep \sim P(X\g \hat{\unobss})$. 

If the p-value
is low, then it implies that the probability of generating a more extreme
dataset than the observed data is small, or in other words, the observed data
itself is considered extreme relative to the model, and thus, the model does
not adequately describe the dataset. In summary, \emph{a low p-value rejects
the hypothesis that the data is being adequately modelled}. The p-value is
usually estimated via an empirical average by generating multiple replicates
$x^\rep_r$, $r=1,\ldots,R$, and evaluating 
\begin{align}
\hat{p}_\textrm{plug-in} =& \frac{1}{R} \sum_{r}
\1(\disc(x^\rep_r, \hat{\unobss}) > \disc(x^\obs, \hat{\unobss})),
\end{align}
where $x^\rep_r$ is a sample from $P(X\g \hat{\unobss})$.

An alternative to point estimation is to consider a Bayesian treatment of the
problem where one can integrate out the contribution of the parameters. The
test statistic can be averaged under either the prior distribution or the
posterior distribution. The \emph{prior-predictive} distribution is defined 
to have the density 
$ p(x^\rep\g \gamma) = \int p(x^\rep\g \unobss)\, p(\unobss \g  \gamma)\, \d\unobss$
where $\gamma$ parameterizes the prior distribution over $\unobs$.
One can generate replicate observations from this distribution, and
compute the \emph{prior predictive p-value} \citep{box-80}
\begin{align}
\pprior = \p(\disc(X^\rep,\unobs) > \disc(x^\obs,\unobs)) 
\approx \; \frac{1}{R}
\sum_{r} \1(\disc(x^\rep_r, \unobss_r) > \disc(x^\obs, \unobss_r))
= \hat{p}_\textrm{prior},
\end{align}
where $(x^\rep,\unobss)_r$ is a sample from $P(X,\unobs\g \gamma)$.
This approach is not reasonable when the prior distribution is improper (cannot
be integrated) or uninformative. Additionally, even if the prior
distribution is informative, one might not generate enough samples to represent
the data distribution well when the parameter space is large. However, notice
that one does not need to fit the model to criticise it.

On the other hand, one can use the posterior distribution $P(\unobs \g
x^\obs)$, and sample from the posterior-predictive distribution with 
density $p(x^\rep\g
x^\obs) = \int p(x^\rep\g \unobss)\, p(\unobss\g  x^\obs)\, \d\unobss$.  The
\emph{posterior predictive p-value} \citep{rubin-84} is then computed as:
\begin{align}
\ppost =  \p(\disc(X^\rep,\unobs) > \disc(x^\obs,\unobs) \g  x^\obs) 
\approx \; \frac{1}{R}
\sum_{r} \1(\disc(x^\rep_r, \unobss_r) > \disc(x^\obs, \unobss_r))
= \hat{p}_\textrm{post},
\end{align}
where $(x^\rep,\unobss)_r$ is a sample from $ P(X^\rep,\unobs\g x^\obs)$,
i.e., by generating samples $\unobss_r$ from the posterior distribution
instead\footnote{Note that the p-value $\ppost(u)=P(D(X^\rep,u) >
(x^\obs,u))$ might be available in closed form depending on the choice of $D$
\citep[Eq.~(8--9)]{Gelman96posteriorpredictive}. Then $\ppost=\frac{1}{R}\sum_r
\ppost(u_r)$ where $u_r$ are posterior samples.}.  
The support of the posterior is usually more concentrated than
prior, and the posterior distribution may be well-defined even if the prior
distribution is improper.

The posterior predictive p-value has been criticised for ``double use'' of data,
once for computing the posterior distribution $P(X^\rep\g x^\obs)$ and
once for computing the discrepancy measure $\disc(x^\obs,\unobs)$
\citep{Bayarri00}.  This means that $\ppost$ does
not have a uniform distribution under the null hypothesis, whereas
$\pprior$ is a valid p-value. $\pplug$ is subject to the same criticism as $\ppost$
since the MLE uses the observed data as well \citep{Bayarri00}.
\citet[\S 7]{lloyd2015statistical} view the different p-values as
arising from ``different null hypotheses and interpretations of the
word `model'\,''. They argued that the posterior predictive and plug-in
p-values are most useful for highly flexible models, as the aim is to
assess the fitted model rather than the whole space of models.
\citet{lloyd2015statistical}
also point out that ``it may be more appropriate to hold out data
and attempt to falsify the null hypothesis that future data will be
generated by the plug-in or posterior distribution'', which is
also in line with the discussion in \cite[\S 2.1]{ohagan-03}.
Further examples of posterior predictive checking can be 
found in \citep{Belin95,GopalanHB15}.

In all of the model criticism described above, a key quantity is the
discrepancy function {$\disc$ used to compare the data and
predictive simulations. We agree with \citet[p.\ 753]{Belin95} who wrote
of the importance of identifying discrepancy functions ``that
would not automatically be well fit by the assumed model", and that
``there is no unique method of Bayesian model monitoring, as there are
an unlimited number of non-sufficient statistics that could be
studied".

\cite{lloyd2015statistical} suggest the Maximum Mean
Discrepancy (\MMD) as a measure of discrepancy between the
observed data and replicates.  The motivation
of using this approach is to maximize the discrepancy over a class of
discrepancy functions rather than choosing only one, i.e.,
\begin{align}
\MMD = \sup_{f \in \F} ( \E_{\data} f(\data) - \E_{X^\rep} f(X^\rep))
\end{align}
where $\F$ is a set of functions.
The function that
maximizes the discrepancy is known as the \emph{witness function}.
When $\F$ is a reproducing kernel Hilbert space (RKHS)
the witness function can be derived in closed form as
\begin{align} \hat{f}(\cdot) =
\frac{1}{|x^\obs|}\sum_{i=1}^{|x^\obs|} \kappa(\cdot,x^\obs_i) -
\frac{1}{|x^\rep|}\sum_{j=1}^{|x^\rep|} \kappa(\cdot, x^\rep_j),
\end{align}
where $\kappa$ is the kernel of the RKHS\@.  This estimation does not work well
in high dimensions, and therefore, the authors suggests reducing the
dimensionality of the observation space before applying this statistic
\cite[p.\,4]{lloyd2015statistical}.

\section{Model Criticism in Latent Space \label{sec:latcritic}}

Recall we have a model $P(X,U\g\gamma)$, with observed variables $X$,
unobserved variables $\unobs$, and known parameters $\gamma$.  In general $U$
may contain latent variables $Z$, parameters $\Theta$, and hyperparameters
$\lambda$.  Our procedure depends on the following two key observations: 
\begin{enumerate}
\item If $x^{\obs}$ is
drawn from the above model, then a sample $u^*$ from $P(U \g x^{\obs}, \gamma)$
is a sample from the prior distribution $P(U \g \gamma)$.  To see why this is true, observe
that a natural way to sample from the joint $P(U,X\g\gamma)$ is to generate 
a sample $u$ from $P(U\g\gamma)$, and then generate a sample $x$ from 
$P(X \g u, \gamma)$ in that order.
However, it is also valid to draw samples from the joint by first sampling 
$x$ from $P(X\g\gamma)$ and then sampling $u$ from $P(U \g x, \gamma)$.  
Thus we have
\begin{stat}
\label{prop:prop} 
If $x^\obs$ is a sample from $P(X\g\gamma)$, then a sample $u^\ast$ from $P(U \g x^\obs, \gamma)$ will be a draw from $P(U\g\gamma)$.
\end{stat}
It is important to clarify what \prop~\ref{prop:prop} is \emph{not} saying. It
is not saying that repeated draws from $P(U\g x^\obs, \gamma)$ will explore the
full prior distribution $P(U\g \gamma)$, but only that it is a valid way to
draw \emph{one} sample from it if $x^{\obs}$ is a draw from the
model\footnote{Throughout this paper, we assume that the prior distribution is
proper, so the respective posterior distribution is well-defined, and that any
MCMC sampler has converged, i.e., the posterior sample is well-behaved.}.
However, testing how well a single draw from a given distribution fits that
distribution is difficult. This brings us to our second observation.

\item If $U$ is a collection of variables, i.e., $U =
(U_1,\ldots,U_K)$, and the prior distribution of $U$ decomposes into
independent draws from the same distribution, e.g., $P(U \g \gamma) =
\prod_{k=1}^K P_u(U_k\g \gamma)$ then it is possible to \emph{aggregate} these
variables together, i.e., instead of testing if $(u_1^\ast,\ldots,u_K^\ast)
$ is a sample from $P(U\g\gamma)$, one can test if $\{u_1^\ast,\ldots,u_K^\ast\}$ is independent and identical draws from the distribution $P_u(\cdot \g
\gamma)$. In other words, rather than testing one sample against a known high
dimensional distribution, one can test if the collection of $K$ samples  are
independent and identical draws from a known lower-dimensional distribution
$P_u$. Thus, we define aggregation as \emph{pooling variables with the same
prior distribution together}, and an \emph{aggregated posterior sample} (APS)
is defined as a set of posterior samples that have been aggregated for
comparison with a specific \emph{reference distribution}. The above
can be generalized to the situation where $U =
(U_1,\ldots,U_K,\theta)$ is a collection of variables and parameters such that
$P(U \g \gamma) = \prod_{k=1}^K P_u(U_k\g \theta)P(\theta\g\gamma)$. Then
$\{u_1^\ast,\ldots,u_K^\ast\}$ can be aggregated and tested against
$P_u(\cdot\g\theta^\ast)$\footnote{Alternatively, $U$ and $\theta$ can be
combined to define a \emph{pivotal quantity} $s$ whose distribution does not
depend on $\theta$ \citep{YuanJohnson12}, and $s(u^\ast,\theta^\ast)$ can be
tested against that distribution.}.
Aggregation can be also extended to the case where $U$ consists of
groups of variables $(U^1,\ldots,U^G)$ where aggregation is performed within
each group $U^g = (U^g_1,\ldots,U^g_{K_g})$ by pooling 
$\{u^{g\ast}_1,\ldots,u^{g\ast}_{K_g}\}$ and comparing against 
$p_{u_g}(\cdot\g u^{- g\ast})$ where $U^{- g}$ denotes all groups except $g$.
We provide more concrete examples of aggregation in
\S\ref{sec:diff-models} and Table~\ref{table:summary}.
\end{enumerate}

We refer to this approach as \emph{aggregated posterior checking} (APC)\@.  We
summarize this approach in Algorithm~\ref{alg:apc}.
Ideas equivalent to \prop~\ref{prop:prop} and the aggregation of
posterior samples can also be found in \citet{YuanJohnson12} \footnote{We had
independently derived the key results.  We thank an anonymous referee for
pointing out the work of \citet{YuanJohnson12}. }, but were applied to the case
where $U$ contains only model parameters, and for hierarchical linear
models. See \S\ref{sec:relwork} for more details on related work.
\begin{algorithm}[h]
\begin{algorithmic}[1]
\caption{Aggregated posterior check}
\label{alg:apc}
\Require Observed data $x^\obs$ 
\Require Bayesian model $P(X\g U,\gamma)P(U\g\gamma)$ with latent variables $U$
\State Generate a posterior sample $u^\ast$ from $P(U\g x^\obs,\gamma)$
\State Generate aggregated posterior sample(s) \Comment{See Table~\ref{table:summary}} 
\State Compare aggregated posterior sample(s) with corresponding reference distribution(s) with appropriate test\\
\Return p-value of the test(s)
\end{algorithmic}
\end{algorithm}

So far, we have addressed the idea of assessing deviations from the prior
distribution and aggregation in the latent space.  However, the same idea can
be applied to the observation space as well, i.e., to the likelihood by
testing if $x^\obs$ is a sample from $P(X\g u^\ast,\gamma)$.  Although it is true that a
discrepancy in the choice of likelihood should be reflected in the posterior
sample, assessing the
discrepancy in the likelihood directly provides better understanding and easier
resolution of the discrepancy.  Notice that, although we make use of $x^\obs$,
our approach is not equivalent to model criticism in the observation space
since we do not compare the observed data $x^\obs$ with replicate observations
$x^\rep_r$, but only investigate the relation between the latent space $u^\ast$
and observation space $x^\obs$. Both methods, however, require generating
posterior samples $u_r$ (for model criticism in the latent space we use $r=1$).

\subsection{Application to different models \label{sec:diff-models}}
We discuss below the application of model criticism in latent space to factor
analysis, linear dynamical systems and Gaussian process regression. These
situations are then demonstrated on real data in \S\ref{sec:results}.

\paragraph{Factor analysis model} 
Consider a  factor analysis model with hyperparameters
$\lambda=\{\tau_\theta,\tau\}$, parameters (loading matrix) $\bfTheta$, latent
variables (factors) $\bfz$ and data $\bfx$. 
Grouping $\bZ = \{ \bfz_i \}_{i=1}^n$ and similarly for $\bX$\footnote{We have omitted the fixed parameters $\gamma$ for simplicity.}, 
\begin{align} \label{eq:phiThetaZXjoint}
p(\lambda, \bfTheta, \bZ, \bX) 
= p(\lambda)\,p(\bfTheta\g \lambda)\, p(\bZ\g \lambda)\, p(\bX\g  \bZ, \bfTheta, \lambda) 
= p(\tau_\theta)\,p(\tau) p(\bfTheta\g \tau_\theta) \prod_{i=1}^n p(\bfz_i)\, p(\bfx_i\g  \bfz_i, \bfTheta, \tau).
\end{align} 
Figure~\ref{fig:mc}d illustrates this model.  In Gaussian factor
analysis, $\bfz \sim {\normal}(\bzero, \inv{\tau_z} \bI)$ and $\bftheta\g \lambda \sim
{\normal}(\bzero, \inv{\tau_\theta} \bI)$ (See Table~\ref{table:dist}).
There is an
identifiability issue in the factor analysis model between $\bfTheta$ and $\bfz$,
which is resolved by fixing the scale of one of the two.  In
Eq.~\eqref{eq:phiThetaZXjoint} the dependence of $\bfz$ on $\lambda$ is taken to be null, i.e., $\tau_z=1$.  
(In the case of example \S\ref{sec:image} we fix the scale of
$\bfTheta$ instead.)
Also, $p(\bfx\g \bfz,\bfTheta,\tau) = \normal(\bfTheta \bfz,
\inv{\tau} \bI)$ and $\tau, \tau_\theta \g \alpha, \beta \sim
\gammarand(\alpha,\beta)$. Thus the fixed parameters
$\gamma=\{\tau_z,\alpha,\beta\}$.

If $\bX^{\obs}$ is drawn from the above model, then a sample $\lambda^*,
\bfTheta^*, \bZ^*$ from $P(\lambda, \bfTheta, \bZ\g \bX^{\obs})$ is a sample from the prior
$P(\lambda) P(\bfTheta\g \lambda)\, P(\bZ\g \lambda)$.  In factor analysis, $\bZ$
decomposes into independent draws from $P(\bfz\g \tau_z)$, and therefore, one
can pool the posterior samples $\bfz^\ast_i$ to assess deviations from
$P(\bfz\g \tau_z)$.  Moreover, each $\bfz_i$ usually decomposes into
independent draws over the different latent dimensions as $\prod_k p(z_{ik}\g
\tau_z)$, one can pool the $z_{ik}^\ast$ to assess deviations from $p(z\g
\tau_z)$.  Similarly, if the prior over the factor loadings matrix $\bfTheta$
decomposes as $p(\bfTheta\g \tau_\theta) = \prod_{kj} p(\theta_{kj}\g
\tau_\theta)$ then one can pool the $\theta_{kj}^\ast$s, and compare with
$p(\theta\g \tau_\theta^\ast)$.  One can also go beyond the marginal $z$ or the
full vector $\bfz$, and assess a subset of the vector such as bivariate
interactions (see \S\ref{sec:image}).

\paragraph{Linear dynamical system}
One can extend the idea of aggregation beyond factor analysis models. For
example, \prop~\ref{prop:prop} holds for general latent variable models with
\emph{repeated structure}. Take, for example, a linear dynamical system model
with a latent Markov chain, so that 
\begin{equation}
p(\bX,\bZ\g U) = p(\bfz_1) p(\bfx_1\g \bfz_1,U)
\prod_{t=2}^T p(\bfz_t\g  \bfz_{t-1},U) p(\bfx_t\g \bfz_t,U)
\end{equation}
where $U$ consists of the system and observation matrices $\bA,\bB$, and
precisions, $\bQ,\bR$.  Then according to \prop~\ref{prop:prop} a sample
$(\bZ^\ast,u^\ast)$ drawn from $P(\bZ,U\g \bX^\obs)$ should be
distributed according to the prior over $(\bZ,U)$.  Although the
$\bfz_t$'s are not independent (due to the Markov chain), we can
consider model criticism for $p(\bfz_t\g \bfz_{t-1})$. For example 
for a system model parameterized as $\bfz_t\g \bfz_{t-1} \sim \bA
\bfz_{t-1} + \bfepsilon_t$ with $\bfepsilon_t \sim \normal(\bzero,\inv{\bQ})$,
violations of the model will show up as deviations of the $\bfepsilon_t^\ast$'s
from $\normal(\bzero,\inv{\bQ^{\ast}})$ (see \S \ref{sec:bee}).  Similarly, for an
observation model parameterized as $\bfx_t\g \bfz_t \sim \bB \bfz_t + \bfpsi_t$
with $\bfpsi_t \sim \normal(\bzero,\inv{\bR})$, violations of the model may also show
up as deviations of the $\bfpsi_t^\ast$'s from $\normal(\bzero,\inv{\bR^{\ast}})$.
See \S\ref{sec:bee}.

\paragraph{Gaussian process regression}
A Gaussian process probabilistic model is defined as:
\begin{subequations}\label{eq:gpeq}
\begin{align}%\tag{Prior}
\vartheta,\zeta,\tau &\sim p(\vartheta)\,p(\zeta)\,p(\tau), \\%\tag{Latent GP}
f(x) &\sim \mathcal{GP}(m(x \g \vartheta),\, \kappa(x,x'\g \zeta)), \\%\tag{Observations}
y_i &\sim \normal(f(x_i),\inv{\tau}) \quad \forall i = 1,\ldots,n,
\end{align}
\end{subequations}
where $m(x\g \vartheta)$ is the mean function parameterized by $\vartheta$,
$\kappa(x,x'\g \zeta)$ is the covariance function (or kernel)
parameterized by $\zeta$, and $\tau$ is the observation noise precision,
see e.g., \citet{Rasmussen2005}. Given observations $\{(\bfx_i,y_i)\}_{i=1}^n$,
$\bfy \sim \normal(\bfm,\bfK)$,
where $\bfy = (y(\bfx_1), \ldots, y(\bfx_n))^\top$, 
$\bfm =( m(\bfx_1\g \vartheta), \ldots, m(\bfx_n\g \vartheta))^\top$
and
$\bfK_{ij} = \kappa(\bfx_i,\bfx_j\g \zeta) + \inv{\tau} \delta(\bfx_i,\bfx_j)$.
Alternatively, considering the eigendecomposition
$\bfK = \bU \bLambda \bU^\top$ where $\bfU =
[\bfu_1,\ldots,\bfu_n]$ is the matrix of the eigenvectors $\bfu_i$s and
$\bLambda$ is the diagonal matrix of the corresponding eigenvalues, i.e.,
$\bLambda_{ii} = \lambda_i$,
\begin{align}\label{eq:gp_unnorm_proj}
\bfc = \bU^\top (\bfy - \bfm) \sim \normal(\bzero,\bLambda).
\end{align}
This implies that, according to the model, the projections $c_i$ of the signal
$\bfy$ on the eigenvector $\bfu_i$ are independent samples from
$\normal(0,\lambda_i)$. Thus, the normalized projections
\begin{align}\label{eq:gp_norm_proj}
\bz = \invhalf{\bLambda}\bU^\top (\bfy - \bfm) \sim \normal(\bzero,\bI)
\end{align}
should be independent samples from $\normal(0,1)$ distribution.
One can thus test the normality of the $z$'s to assess the goodness of
fit. However, note that if the $i$th eigenvalue of $\bfK$ is much smaller
than the noise variance $\inv{\tau}$, then this $z_i$ is dominated by the white noise
contribution. Thus we only include $z$'s corresponding to eigenvalues
with $\lambda_i > 2 \inv{\tau}$ to assess the fit of the GP
model\footnote{The factor of 2 on the RHS of the inequality is
included because the $\lambda_i$'s are shifted by $\inv{\tau}$ by
definition.}. See \S \ref{sec:co2}.

\subsection{Related Work \label{sec:relwork}}

\citet{Cook06} consider the situation with (in our notation) a prior $p(u)$ on
the parameters of the model, and data $p(x\g u)$. They then assume that
specific parameters $u^0$ are drawn from the prior, then data $x^{\mathrm{obs}}$ drawn from $P(X\g u^0)$. They then consider samples $u^1, u^2, \ldots, u^L$ drawn from
$P(U\g x^\obs)$, and comment (in the caption of their Figure 1, translating to
our notation) that ``$(x^\obs,u^{\ell})$ should look like a draw from $P(X,U)$
for $\ell = 0, 1, \ldots, L$''. They then use the `reverse' of
\prop~\ref{prop:prop} to validate the correctness of posterior samples
generated by a statistical software, by comparing $u^0$ with $u^1, u^2, \ldots,
u^L$. Their recommended method for this is to calculate posterior quantiles for
each scalar parameter; if the software is working correctly then the posterior
quantiles are uniformly distributed. Although they share with us the
observation that $(x^{\mathrm{obs}},u^{\ell})$ should look like a draw from
$P(X,U)$, this is used to answer a totally different question. Also, they do
not discuss the inclusion of latent variables in the model.

\citet{johnson2007} and later \cite{YuanJohnson12} also consider a
model with parameters $U$ and data drawn from $P(X\g U)$. Their
interest is in the use of pivotal quantity $d(x, u)$ that has a known
and invariant sampling distribution when data $x^\obs$ are generated
from a model with data-generating parameters $u^0$. Then
\cite{YuanJohnson12} show that if the $d(X, u^0)$ is a pivotal
quantity distributed according to $F$, then $d(X, u^{\ell})$ is also
distributed according to $F$, if $ u^{\ell}$ is drawn from the
posterior on $ U$ given $x^\obs$. The result of \cite{YuanJohnson12}
extends earlier work by \citet{johnson2007} to the case where $d(x,
u)$ does not depend on the data $x$---for example this situation can
arise in a Bayesian hierarchical linear regression model, when
considering the second level where parameters for individual units are
generated from a hyperprior.

Regression diagnostics is a well-explored example of model criticism.  Existing
approaches assess certain statistical assumptions made during modelling, e.g.,
if the residuals follow a normal distribution with zero mean, (e.g., using a
Q-Q plot \citep{Wilk68}), if the residuals are homoscedastic, (e.g., using the
Breusch--Pagan \citep{Breusch79} or White test \citep{White80}) or if the
successive residual terms are uncorrelated (e.g., using the Durbin--Watson test
\citep{Durbin50}). Regression diagnostics can be seen as a special case of
model criticism in the latent space since residuals are representatives of
\emph{errors}, which are latent variables of the model. However, our 
methods are also applicable to more complex models.

\cite{meulders-etal-98} consider a factor analysis model
for binary data, using (in our notation) $\betarand(2,2)$ priors on
$\bZ$ and $\bfTheta$. They carry out posterior sampling using
block Gibbs sampling for $\bZ$ and $\bfTheta$ and compare histograms
of these variables against the prior.  Discrepancies between the
prior and histograms of the sampled aggregated posterior led to
model extension, expanding the model to use a mixture of two
beta distributions for the parameters. However, the authors do not
explain the basis for carrying out this check (cf.\ 
\prop~\ref{prop:prop}). 

\cite{buccigrossi-simoncelli-99} consider the posterior distribution
of wavelet coefficients (analogous to $\bfz$ in the factor analysis
model) in response to image patches. By considering the
distribution of a bivariate aggregated posterior, they show that this
is not equal to the product of the marginals, but
exhibits variance correlations. (This is shown by introducing a
``bowtie plot'' showing the conditional histogram of $z_2$ given
$z_1$.) This work is a nice example of how the failure of a diagnostic
test can give rise to an extended model (see \S\ref{sec:image}).

\citet[\S 3]{ohagan-03} considered model criticism tools that can be applied at
each node of a graphical model (and of course latent variables can be
considered as such).
\citet[\S3.1][]{ohagan-03} discussed the idea of residual testing at different
levels of a hierarchical model as well as a generic probabilistic model.
He suggested checking if a node in a probabilistic model is misbehaving by
comparing the posterior samples at that node to prior distribution.
\citet[\S3.2][]{ohagan-03} also emphasized that conflict can arise between the
different sources of information about a variable at a particular node,
arising from contributions from each neighbouring node in the graph. However,
he did not suggest using the aggregated posterior to assess
goodness of fit, but considered the posterior at each node separately.

\cite{Tang12} introduce the concept of the ``aggregated posterior'' as
applied to deep mixtures of factor analysers (MFA) model.
Consider the situation as above but where $\bfTheta$ is estimated by
maximum likelihood, so it is the posterior over $Z$ that is of
interest. Thus
$p(\bX,\bZ\g \bfTheta) = \prod_{i=1}^n p(\bfz_i)\, p(\bfx_i\g \bfz_i,\bfTheta).$
Under this model we also have that
$p(\bfz) = \int p_{\bfTheta}(\bfz\g \bfx)\, p_{\bfTheta}(\bfx) d\bfx$
where the $\bfTheta$ subscript denotes that both
$p_{\bfTheta}(\bfz\g \bfx)$ and $p_{\bfTheta}(\bfx)$ correspond to
distributions under the model. \citet[][p3]{Tang12} define the aggregated
posterior as ``the empirical average over the data of the posteriors
over the factors'', i.e.,
\begin{equation} \label{eq:aggpost}
\tilde{p}(\bfz) = \frac{1}{n}\sum_{i=1}^n p_{\bfTheta}(\bfz\g \bfx_i^\obs),
\end{equation}
where the integral wrt $p_\bfTheta(\bfx)$ has been replaced by the empirical
average over samples.
If the data distribution $p(\bfx)$ is equal to the model
distribution $p_{\bfTheta}(\bfx)$ then $\tilde{p}(\bfz)$ should agree
with $p(\bfz)$. However, differences between $p(\bfx)$ and
$p_{\bfTheta}(\bfx)$ will manifest as
differences between the two respective distributions in the latent space.

In practice, however, one does not explicitly construct the
aggregated posterior Eq.~\eqref{eq:aggpost}
since it is only asymptotically equal to the prior. Instead \cite{Tang12}
compare a collection of $n$ samples
$\b{z}_i^\ast$ from $p_\bfTheta(\b{z}\g \b{x}_i^\obs)$ for $i = 1, \ldots, n$ to
$p(\b{z})$. This is a valid approach since if
$\{\bfx^\obs_1,\ldots,\bfx^\obs_n\}$ follow the distribution
$p_\bfTheta(\bfx)$, then $\{\b{z}_1^\ast,\ldots,\b{z}_n^\ast\}$ follow
the distribution $p(\b{z})$ as we show in \prop~\ref{prop:prop}.
Additionally, as $\bfTheta$ is not known in practice, \cite{Tang12}
replace $\bfTheta$ with maximum likelihood estimate $\hat{\bfTheta}$ in
the definition of aggregated posterior Eq.~\eqref{eq:aggpost}.
In \prop~\ref{prop:prop} we extend this idea to a Bayesian
setting where $\bfTheta$ and $\lambda$ are not fixed parameters
but latent variables themselves.

\cite{Tang12} started
with a simple mixture of factor analysers (MFA), and observed that
the aggregated posterior for a latent component often doesn't match the $\normal(\bzero,\bI)$ prior.
By replacing the prior for a component with
another MFA model, they constructed a deep MFA model. The
idea of the aggregated posterior (although not the name) can be traced
back e.g.\ to \citet{hinton-osindero-teh-06}, where in deep belief
nets the idea was that the posterior distribution of the latents of a
restricted Boltzmann machine (RBM) could be modelled by another RBM\@.

\section{Examples \label{sec:results}}
In this section, we provide three examples of model criticism and extension in
the latent space.  First, we explore a factor analysis model in the context of
image compression (\S\ref{sec:image}).  The objective of this example is to
show how the model can be criticised in the latent space as well as in the
observation space. Our analysis leads to changing the latent distribution from
a single Gaussian to a scale mixture of Gaussians, which captures both the
marginal and the joint structure of the latent space, and improves the model
in the observation space as well.

Next, we explore a linear dynamical system model (\S\ref{sec:bee}) in the
context of modelling time series. We show that model criticism in latent space
allows us to interrogate not only the standard ``innovations'' (defined in
\eqref{eq:bee_agg2}), but also the latent residuals (defined in
\eqref{eq:bee_agg1}).

Finally, we explore a Gaussian process model (\S\ref{sec:co2}) in the
context of modelling time series. The objective of this example is to
show when model criticism in the latent space can be a natural
choice whereas model criticism in the observation space can be
difficult. Our analysis leads to changing the covariance
function from squared exponential to a combination of periodic and
squared exponential kernels.

We implemented all models (except the Gaussian process model)
in JAGS \citep{Plummer03jags}, keeping a single
sample in the MCMC run after discarding a burn-in of 1000 samples (10,000
samples for \S\ref{sec:bee}). Note that for model criticism in the latent
space, we need \emph{only a single sample}.
We summarize the aggregation process and corresponding reference
distributions used in this section in Table~\ref{table:summary}.
\begin{table}[h]
\centering
\scriptsize
\begin{tabular}{ccccc}
\toprule
Model & $x^\obs$ & $U$ & APS & reference distribution \\
\midrule
MF~\eqref{eq:imagepatch} & 
$\{\bx_i^\obs\}_{i=1}^{n}$ & 
\makecell{$\{\bz_i\}_{i=1}^n, \bfTheta, \bfb, \tau, \tau_z$ for~\eqref{eq:eq:Gaussian}-\eqref{eq:eq:Laplace} \\
$\{\bz_i\}_{i=1}^n, \bfTheta, \bfb, \tau, \bpi, \bftau$ for~\eqref{eq:eq:scalemxgauss} } & 
\makecell{$\{z^\ast_{ki}\}_{i=1,k=1}^{n,K}$, and \\
$\{(z^\ast_{k_1i},z^\ast_{k_2i})\}_{i=1,\,k_1,k_2=1,k_1\neq k_2}^{n,K}$} & 
\makecell{
$\normal(0,\inv{\tau_z^{\ast}})$, and \\
$\normal(0,\inv{\tau_z^{\ast}} \bI_2)$ for~\eqref{eq:eq:Gaussian} \\
$\doubleexp(0,\tau^\ast_z)$, and \\
$\doubleexp(z_1;0,\tau^\ast_z)\,\doubleexp(z_2;0,\tau^\ast_z)$
for~\eqref{eq:eq:Laplace} \\
$\sum_m \pi^\ast_m\normal(0,\inv{\tau^{\ast}_m})$, and \\
$\sum_m \pi^\ast_m\normal(\bzero,\inv{\tau^{\ast}_m} \bI_2)$ for~\eqref{eq:eq:scalemxgauss} } \\
\hline
\makecell{LDS\\\eqref{eq:slds},~\eqref{eq:slds_prior}} & 
\makecell{$\{(x,y,$\\$\cos(\nu),\sin(\nu))^\obs_t\}_{t=1}^n$} & 
\makecell{$\{s_t\}_{t=1}^n,\{\bfz_t\}_{t=1}^n,\bA^{(1)},\ldots,$ \\ $\bA^{(S)},\bQ^{(1)},\ldots,\bQ^{(S)},\bB,\bR$} & 
\makecell{
$\{\tilde{z}^\ast_{kt}\}_{t=2}^n$ from \eqref{eq:bee_agg1} and \\ 
$\{\tilde{x}^\ast_{jt}\}_{t=2}^n$ from \eqref{eq:bee_agg2} $\forall\,j,k$\\
$\{(\tilde{z}^\ast_{kt}, \tilde{z}^\ast_{k(t+1)})\}_{t=2}^{n-1}$ from \eqref{eq:bee_agg1} and \\ 
$\{(\tilde{x}^\ast_{kt}, \tilde{x}^\ast_{k(t+1)})\}_{t=2}^{n-1}$ from \eqref{eq:bee_agg2} $\forall\,j,k$}& 
\makecell{
$\normal(0, 1)$ \\
$\normal(0, 1)$ \\
$\normal(\bzero, \bI_2)$ \\
$\normal(\bzero, \bI_2)$} \\ 
\hline
GP~\eqref{eq:gpeq} & 
$\{(x_i,y_i)^\obs\}_{i=1}^n$ & 
\makecell{ $\sigma^2_f,l,\tau$ for~\eqref{eq:sekernel}\\
$\sigma^2_f,p,l_p,l_d$ for~\eqref{eq:sekernel} and~\eqref{eq:pekernel}\\
$\sigma^2_f,f,l_p,l_d,\sigma^2_{fs},l_s,\sigma^2_{fl},l_l $ \\
for~\eqref{eq:sekernel} (large and small) and~\eqref{eq:pekernel}
} & 
$\{z^\ast_i\}_{i=1}^n$ from~\eqref{eq:standproj} & 
$\normal(0, 1)$ \\
\bottomrule
\end{tabular}
\caption{The table summarizes observed data $x^\obs$, unknown variables $U$,
aggregated posterior sample(s) (APS), and corresponding reference
distribution(s) (as elaborated in Algorithm~\ref{alg:apc}) for three models
discussed in \S\ref{sec:results}, and different scenarios within each model.}
\label{table:summary}
\end{table}

\begin{figure*}[t]
\centering
\begin{minipage}[l]{0.32\textwidth}
\centering\footnotesize{(a) \ref{tag:eq:Gaussian} \eqref{eq:eq:Gaussian}}
\includegraphics[width=\textwidth]{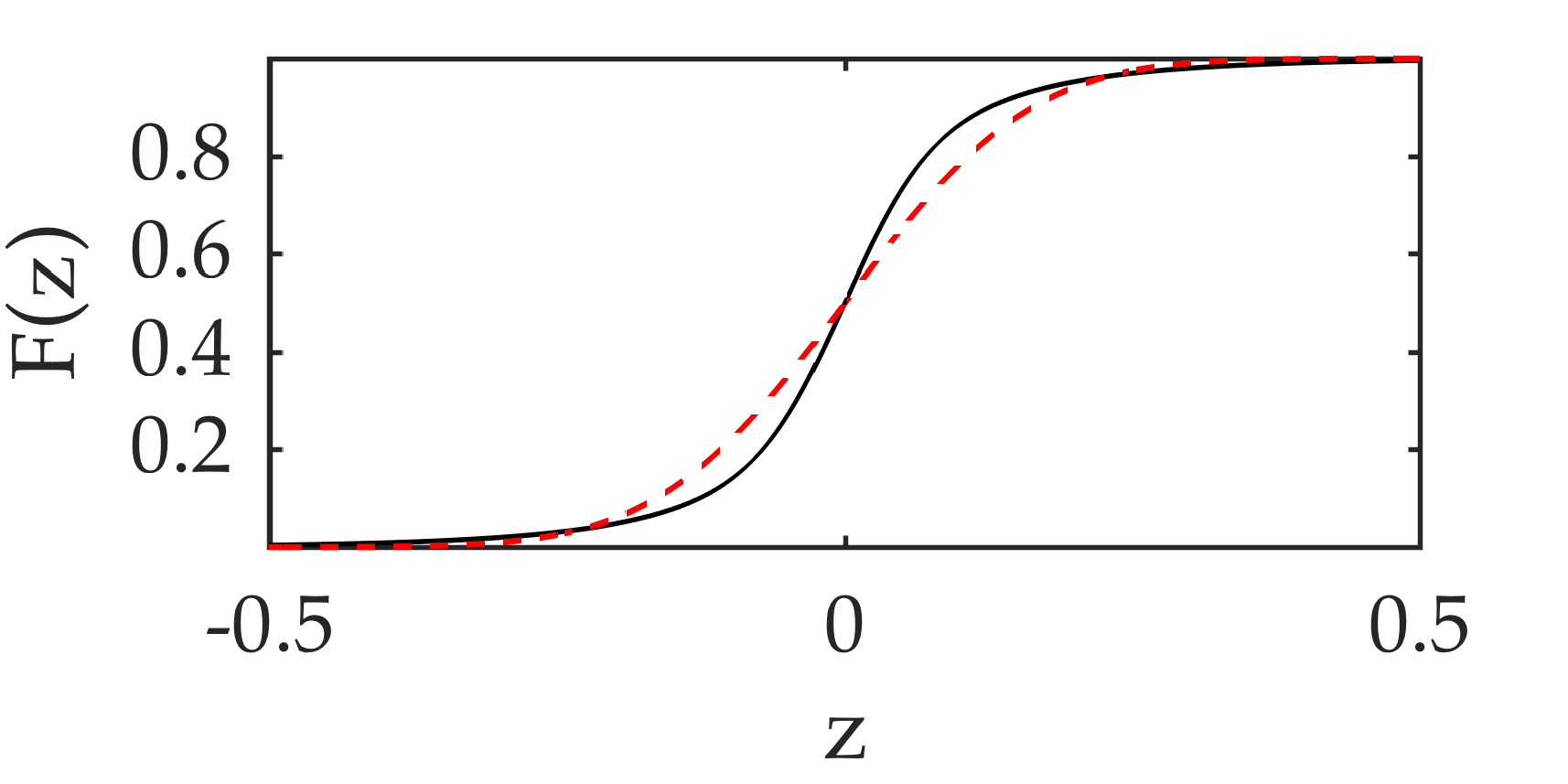}\\
\includegraphics[width=\textwidth]{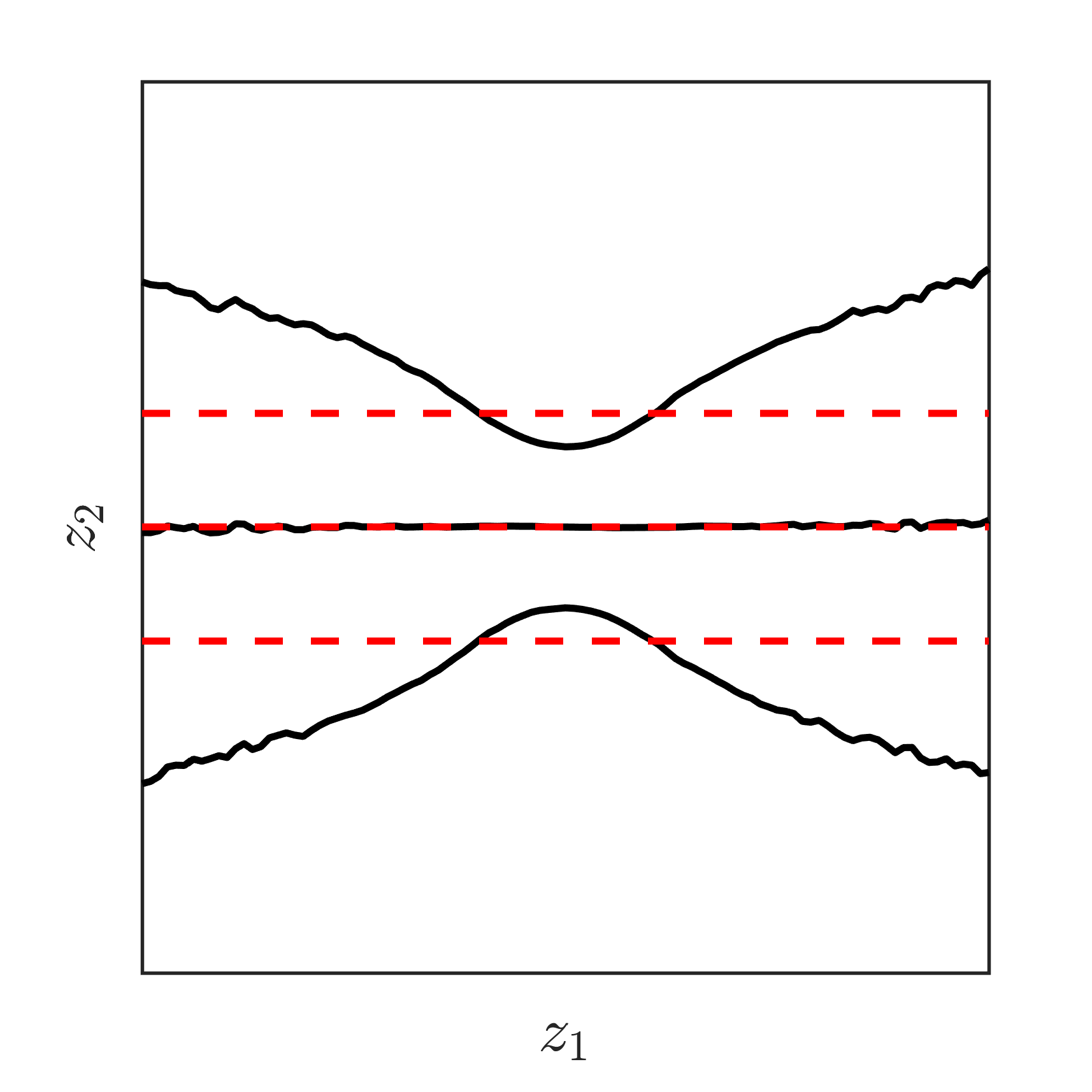}\\
\includegraphics[width=0.8\textwidth]{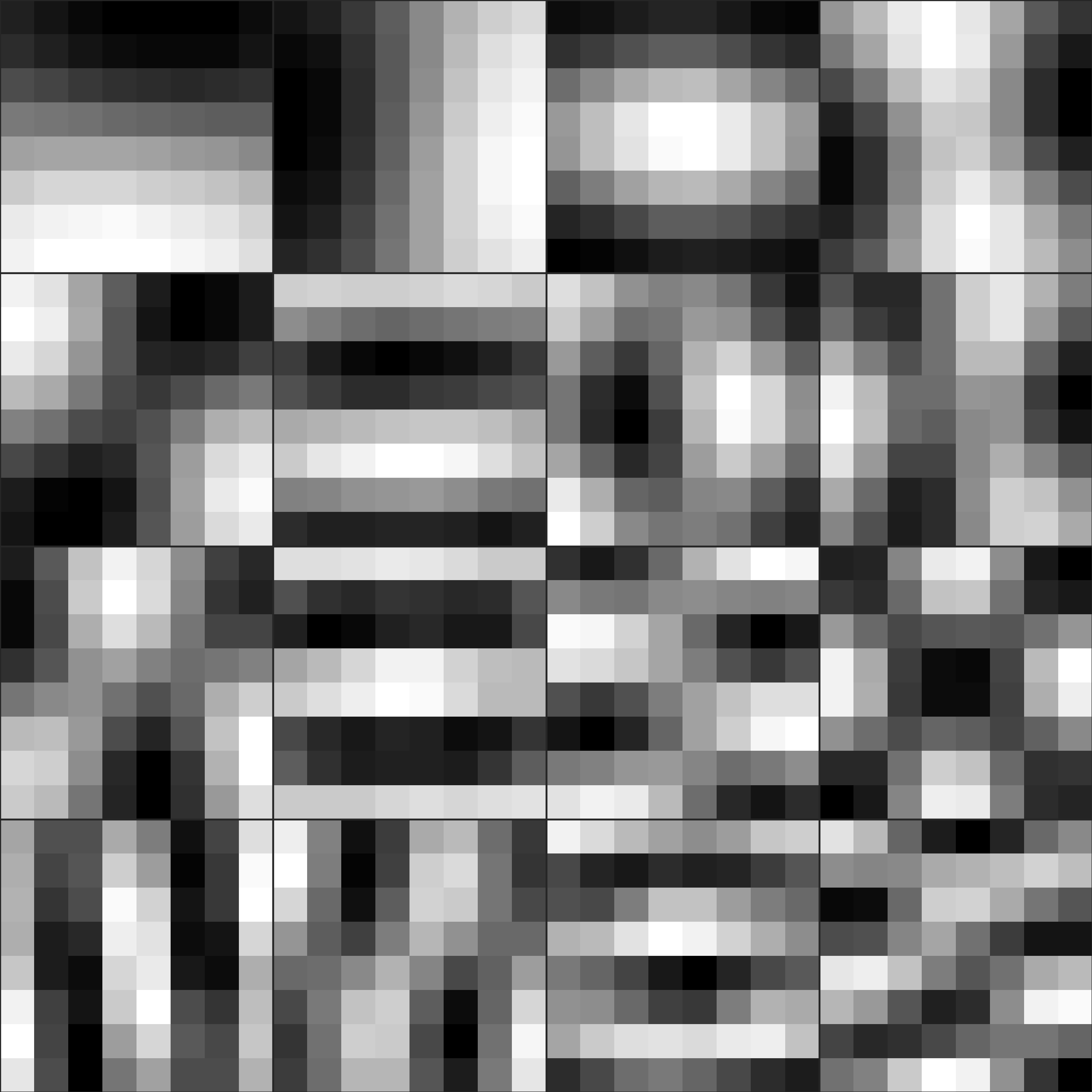}
\end{minipage}
\begin{minipage}[c]{0.32\textwidth}
\centering\footnotesize{(b) \ref{tag:eq:Laplace} \eqref{eq:eq:Laplace}}
\includegraphics[width=\textwidth]{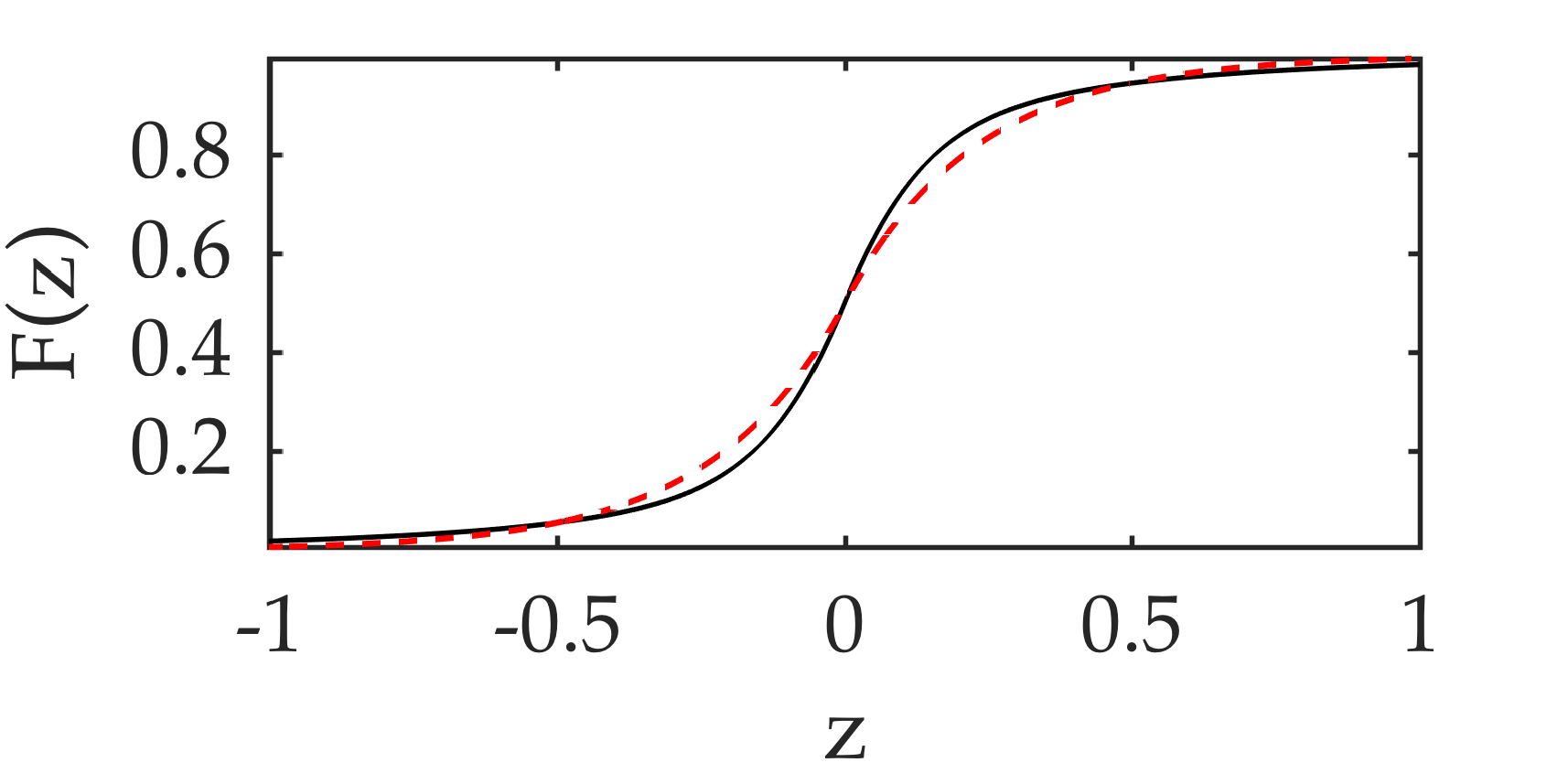}\\
\includegraphics[width=\textwidth]{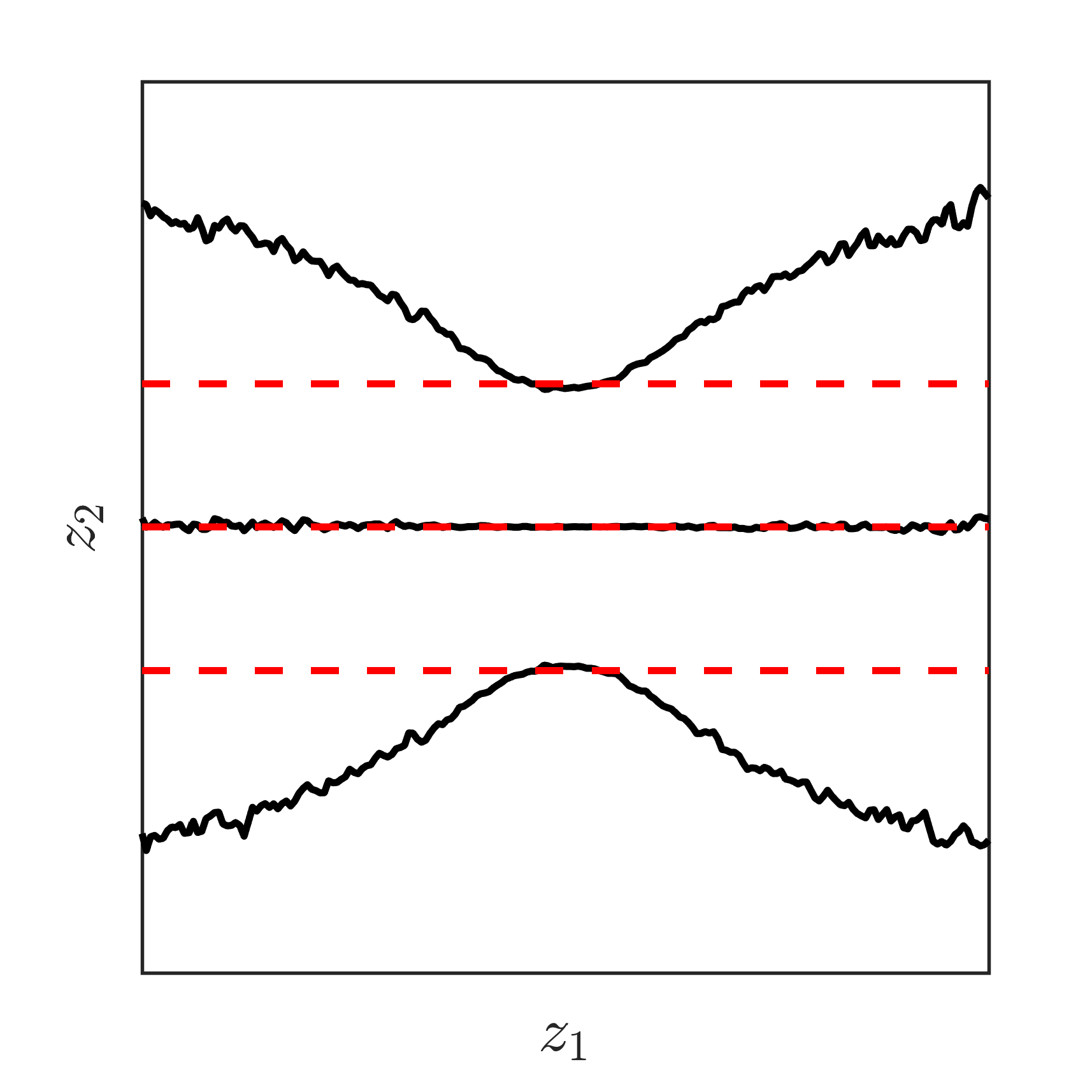}\\\,
\includegraphics[width=0.8\textwidth]{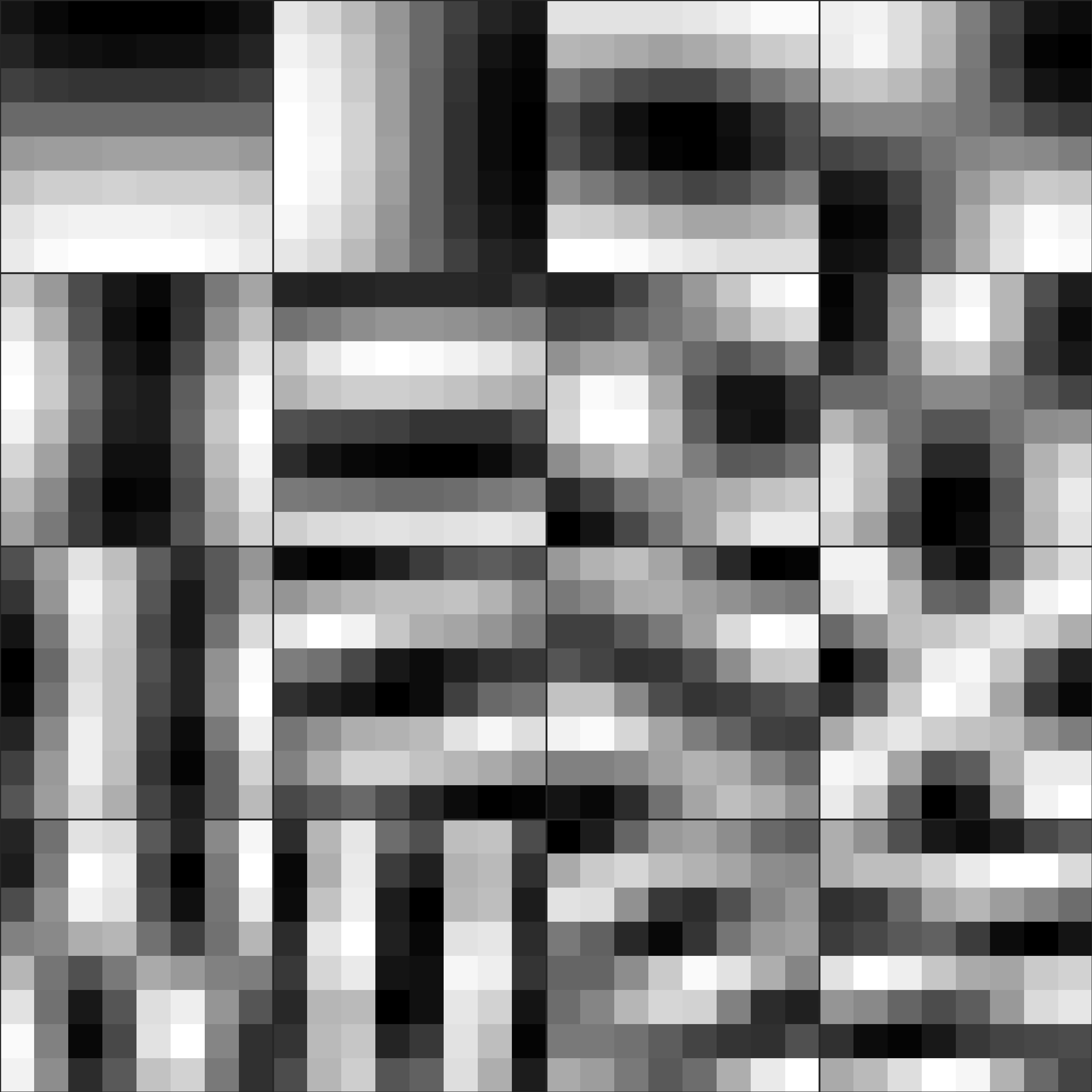}
\end{minipage}
\begin{minipage}[r]{0.32\textwidth}
\centering\footnotesize{(c) \ref{tag:eq:scalemxgauss} \eqref{eq:eq:scalemxgauss}}
\includegraphics[width=\textwidth]{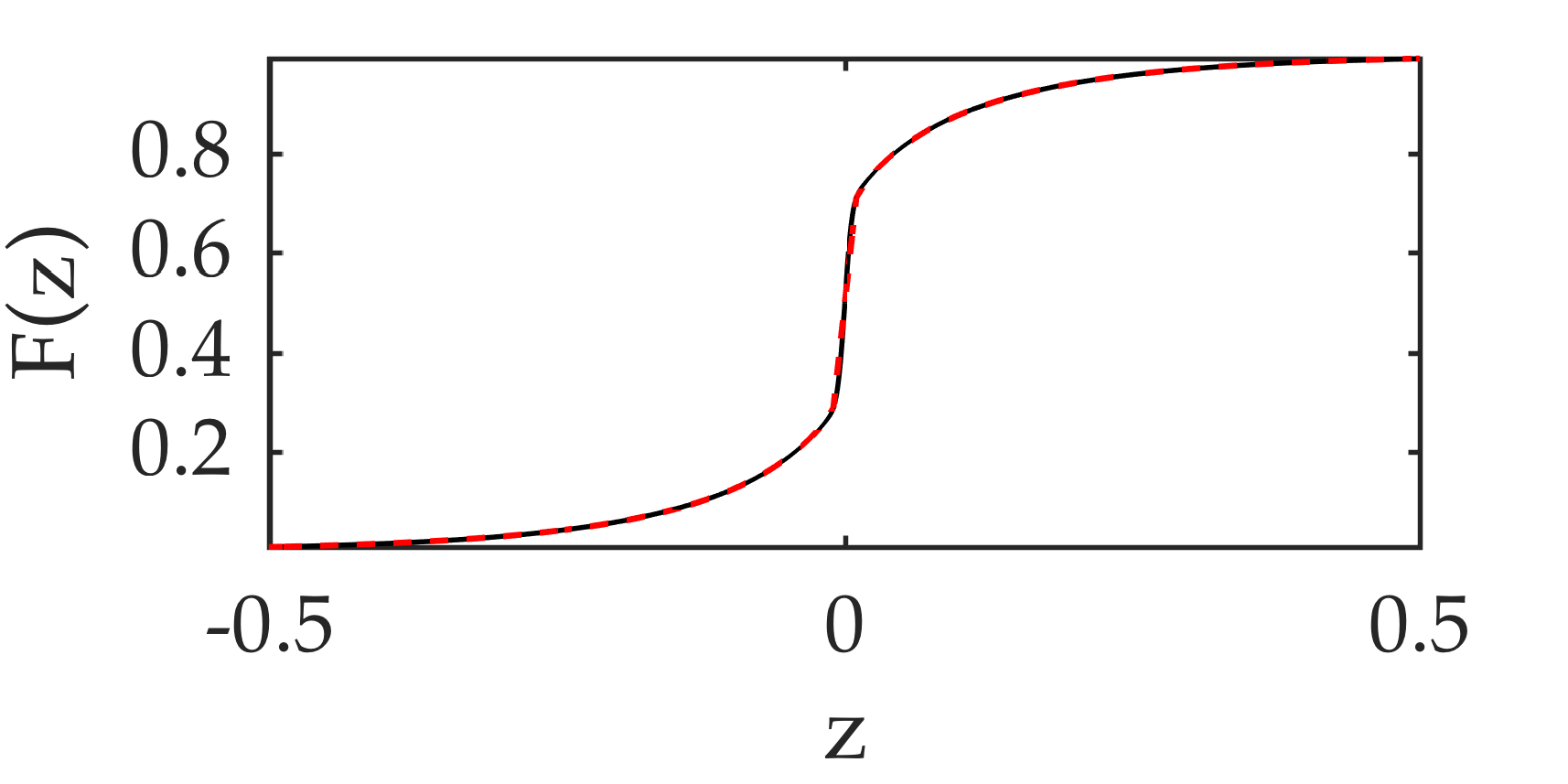}\\
\includegraphics[width=\textwidth]{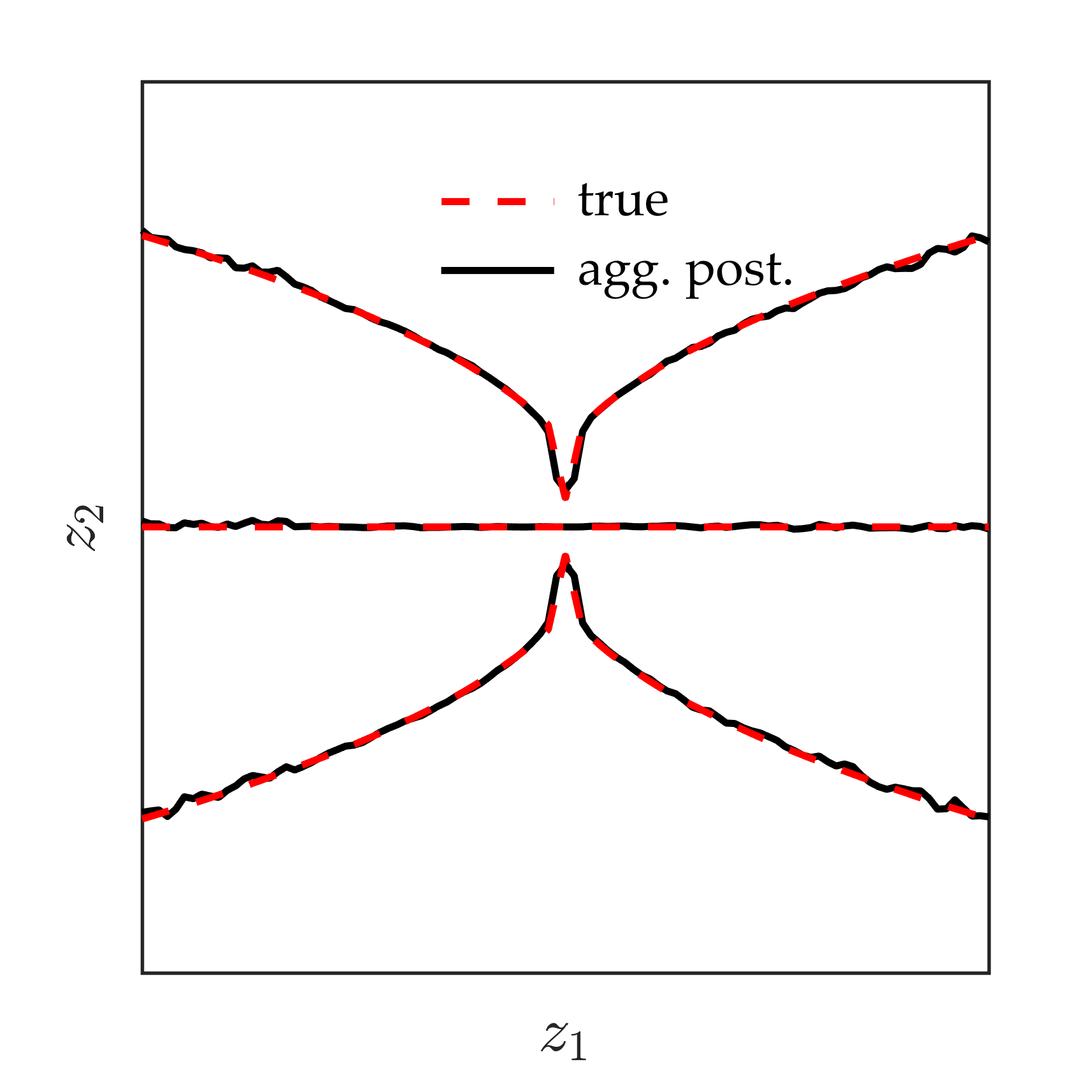}\\\,
\includegraphics[width=0.8\textwidth]{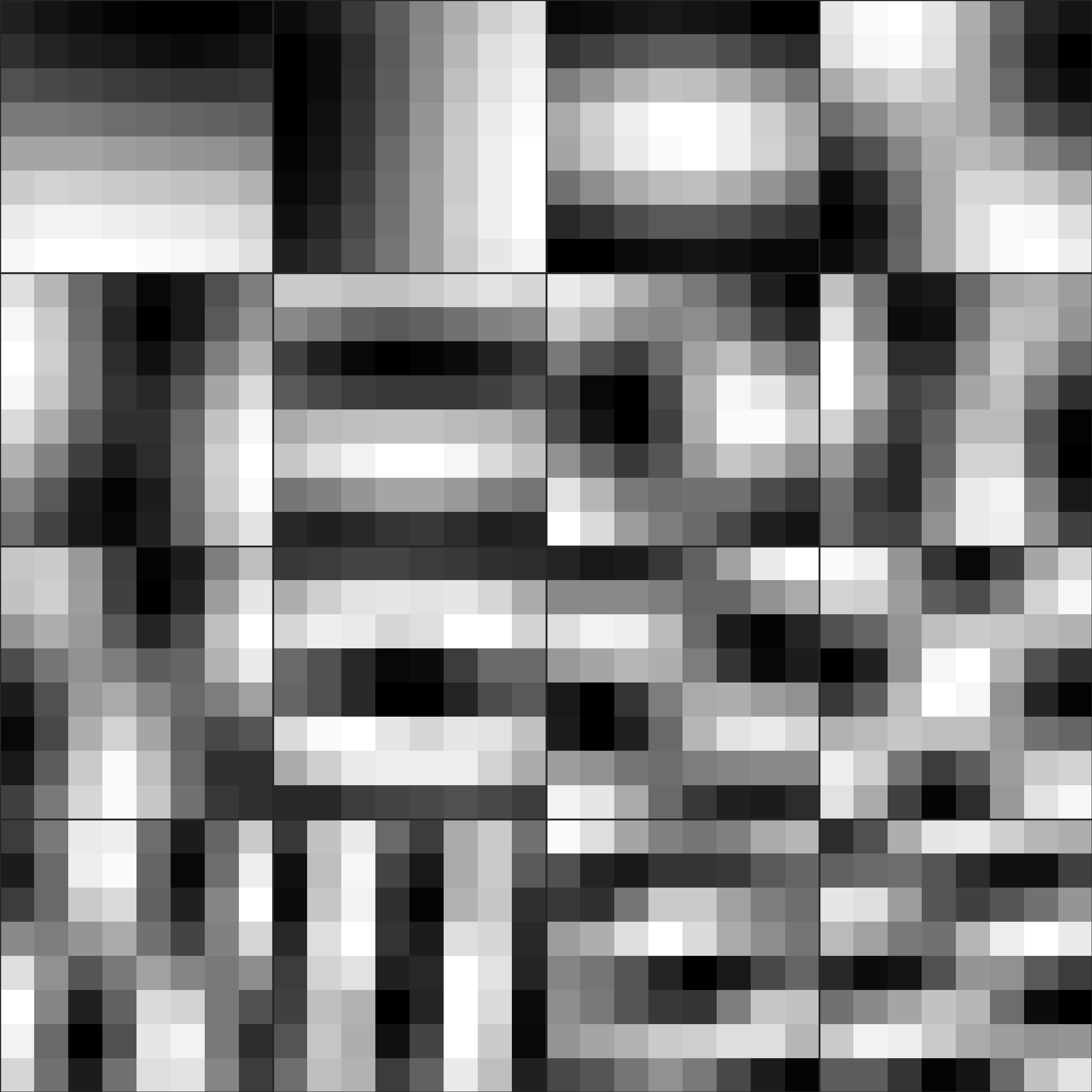}
\end{minipage}
\caption{Aggregated posterior [\textrm{agg.~post}] and associated prior
at the latent node for (left to right) the Gaussian,
Laplace and Scale mixture of Gaussian models of image patches.  \textbf{Top}: ECDF of
aggregated posterior samples
($16\times 50,000$ samples) and CDF of respective prior.
\textbf{Middle}: conditional mean and standard deviation of the bivariate
aggregated posterior samples ($16\times 15 \times 50,000 \div 2$ samples, over 100
bins) and respective prior distribution. \textbf{Bottom}: eigenvectors of
respective loading matrices.}
\label{fig:image}
\end{figure*}

\subsection{Image patch data \label{sec:image}}
The Berkeley Segmentation Database \citep{MartinFTM01}
consists of 200 training images.
Following \citet{Zoran12}, we convert the images to greyscale, extract
$8 \times 8$ randomly located patches, and remove DC components from
all image patches\footnote{https://people.csail.mit.edu/danielzoran/NIPSGMM.zip}. We extract 50,000 image patches, and fit different
matrix factorization models of the form:
\begin{subequations}\label{eq:imagepatch}
\begin{align}
\b{b} &\sim \normal(\bzero,\bI),\,
\tau \sim \gammarand(\alpha,\beta),\,
\theta_{jk} \sim \normal(0,1) \\
\b{z}_i &\sim \mathrm{LatentDist},\,
\b{x}_i \sim \normal(\bfTheta\b{z}_i + \b{b},\, \inv{\tau} \bI)
\end{align}
\end{subequations}
with $K=16$ latent dimensions ($>82\%$ explained variance in PCA)\@.
Previously \citet{Zoran12} used full covariance zero-mean Gaussians
($\tau\!=\!0$, $K\!=\!64$, $\b{b}\!=\!\mathbf{0}$). We set
$\alpha\eq \beta\eq 0.001$.

We start by assuming a Gaussian distribution for the latent model
\begin{align}
\tau_z \sim \gammarand(\alpha,\beta), \, z \sim \normal(0, \inv{\tau_z}), \owntag[eq:Gaussian]{Gaussian}
\end{align}
and generate a sample
$(\bZ^\ast,\bfTheta^\ast,\bfb^\ast,\tau^\ast,\tau_z^\ast)$ from the
posterior.  To criticise the model, we aggregate the univariate
posterior samples $\{z^\ast_{ki}\}\,\forall k,i$, and
bivariate samples $\{(z^\ast_{ki_1},z^\ast_{ki_2})\}\,\forall k,i_1\neq i_2$. If the
observed data follows the model, then the
distributions of the corresponding APSs
are univariate normal $\normal(0,\inv{\tau_z^{\ast}})$
and bivariate normal
$\normal(\bzero,\inv{\tau_z^{\ast}} \bI_2)$ respectively. We observe that neither of the
APSs follow the expected prior
distribution. Also, the marginal distribution is more concentrated
around zero than the expected distribution, whereas the joint
distribution shows heteroscedasticity (Figure~\ref{fig:image}, left
column) which is inconsistent with the factorized bivariate
normal prior.

An alternative latent variable model for the factor analysis model is
(See Table~\ref{table:dist})
\begin{align*} 
z &\sim \doubleexp(0,\tau_z). \owntag[eq:Laplace]{Laplace}
\end{align*} 
We use the same aggregation strategy as before, and compare the empirical
distributions with the univariate distribution $\doubleexp(0,\tau^\ast_z)$ and
bivariate distribution
$\doubleexp(z_1;0,\tau^\ast_z)\,\doubleexp(z_2;0,\tau^\ast_z)$
respectively.  We observe similar characteristics in the aggregated posterior
as before (Figure~\ref{fig:image}, middle column).

To accommodate these observations,
we allow a scale mixture of Gaussian distributions as used by
\citet{Wainwright99b} with 8 components for the latent
variable
\begin{align}
\pi \sim \Dirichlet(\b{1}), \, \tau_m \sim \gammarand(\alpha,\beta),\,
\b{z} \sim \sum_{m=1}^8 \pi_m \normal(\bzero,\inv{\tau_m} \bI) 
\owntag[eq:scalemxgauss]{Scale Mixture of Gaussians}.
\end{align}
We generate sample $(\pi^\ast_1,\ldots,\pi^\ast_8,
\tau^\ast_1,\ldots,\tau^\ast_8)$ as well. We
assess the same aggregated distributions as before and
compare them with
$\sum_m \pi^\ast_m\normal(0,\inv{\tau^{\ast}_m})$, and
$\sum_m \pi^\ast_m\normal(\bzero,\inv{\tau^{\ast}_m} \bI_2)$ respectively.
We observe that the empirical marginal distribution follows the mixture
distribution well,
although a KS test rejects the hypothesis that the
aggregated posterior follows the mixture distribution.
Additionally, the joint distribution captures the
heteroscedasticity in the latent space (Figure~\ref{fig:image}, right
column).

We also show the eigenvectors of the corresponding
loading matrix for each of the three cases (Figure \ref{fig:image} bottom
row). We show the eigenvectors rather than the loading matrix themselves since
for the Gaussian and Gaussian scale mixture, the columns of the loading matrix
may not correspond to any particular pattern due to rotational invariance. We
observe that all three loading matrices span a similar space.
\begin{figure*}[t]
\centering
\includegraphics{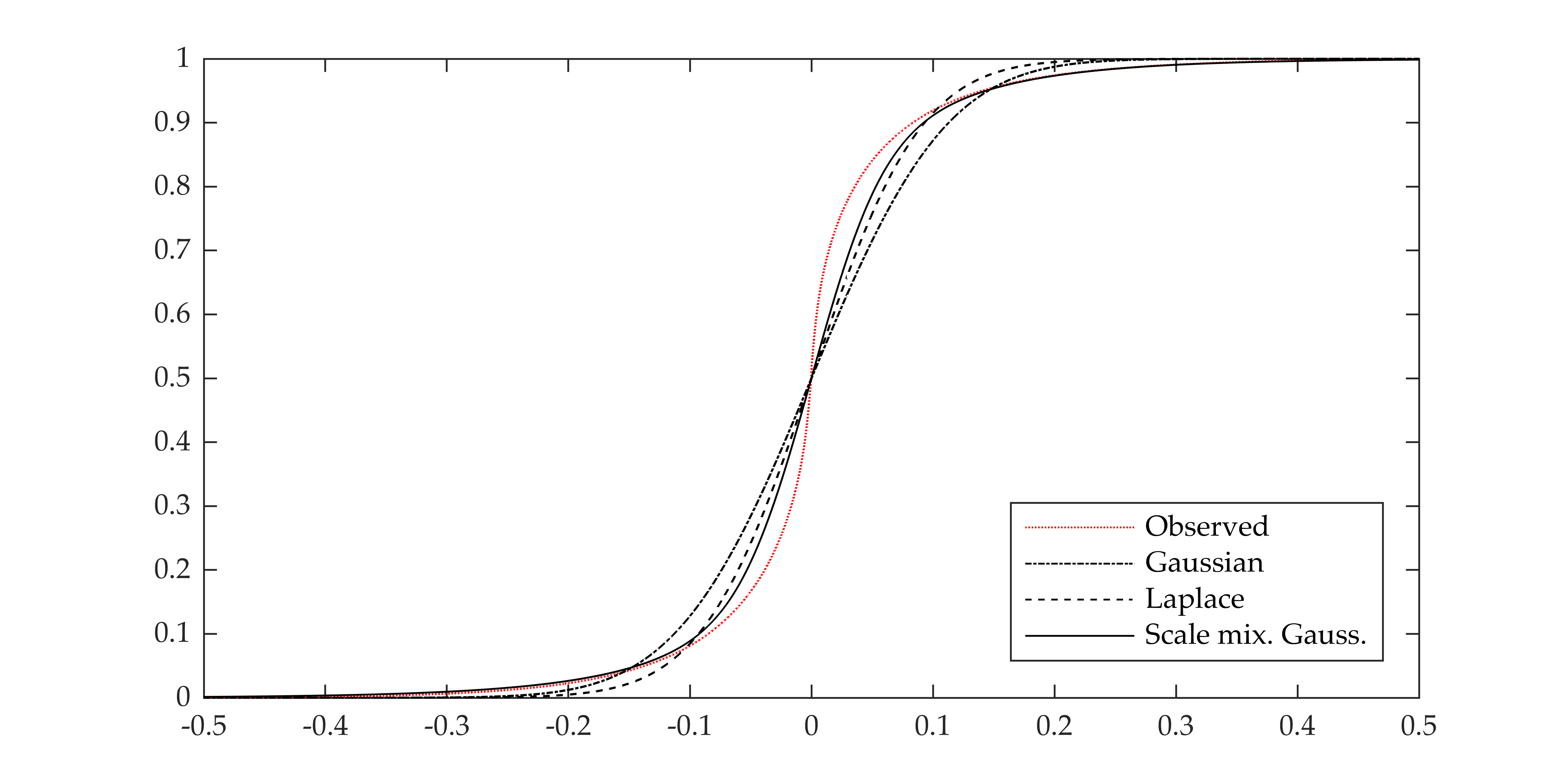}
\caption{ Empirical cumulative distribution functions of raw pixel values on
observed and replicate data for varying different distributions.}
\label{fig:imagePatchGradient}
\end{figure*}

The matrix factorization model can be criticised in the observation space
with established image statistics as a discrepancy measure. This, however,
requires generating replicate data of the same size as the observed data, which
in this case is computationally extensive since $\bX^\obs \in \reals^{64 \times
50,000}$.  To avoid generating multiple replicates, i.e., matrices $\bX^\rep_r
\in \reals^{64 \times 50,000}$, we only generate a single replicate for each
latent distribution choice and compare them the observed data.

For all three cases, i.e.,~\ref{tag:eq:Gaussian}, \ref{tag:eq:Laplace}, and
\ref{tag:eq:scalemxgauss}, we generate latent samples $\bz^\rep_i$ from the
fitted parameters $\tau_z^\ast$ (and $\bftau,\bpi^\ast$ for \ref{tag:eq:scalemxgauss}).
We use the rest of the fitted parameters, i.e., $\bfTheta^\ast$, $\tau^\ast$, and
$b^\ast$ to generate samples $\bx^\rep_i$ from $\bz^\rep_i$. We generate
50,000, $8 \times 8$ replicate image patches, and compare the observed and
replicate data in terms of the distribution of raw pixel values.  We show the
results in Figure~\ref{fig:imagePatchGradient}. We observe that the
distribution of the image pixel values in the replicate data follows the
observed data more closely for \ref{tag:eq:scalemxgauss} than the other latent
distributions. However, it is not a perfect fit, and that tells us that this
model can improved further; potentially by increasing $K$, and varying the
noise characteristics such as using a full diagonal covariance.

\begin{figure*}[t]
\centering
\includegraphics{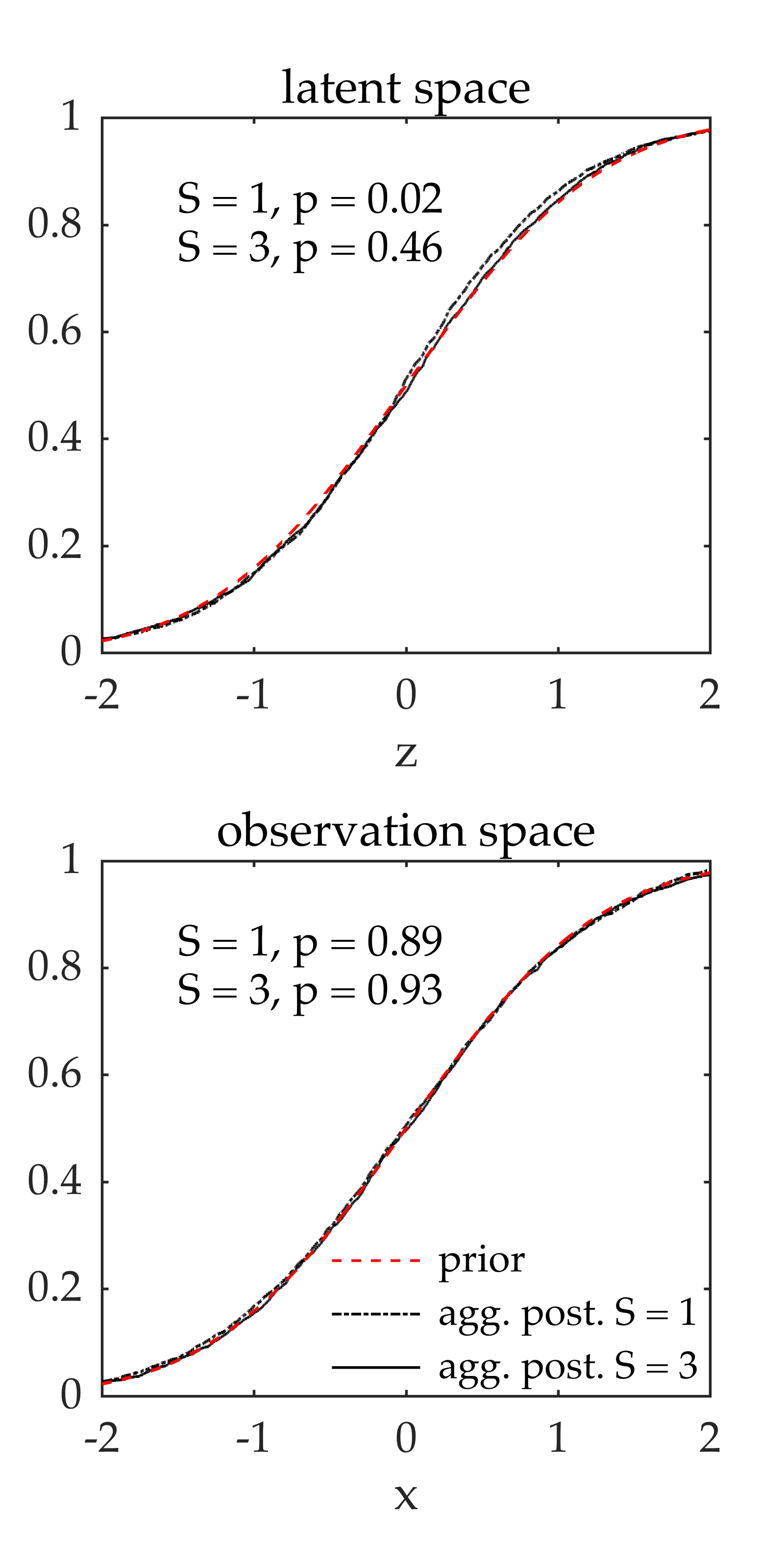}
\includegraphics{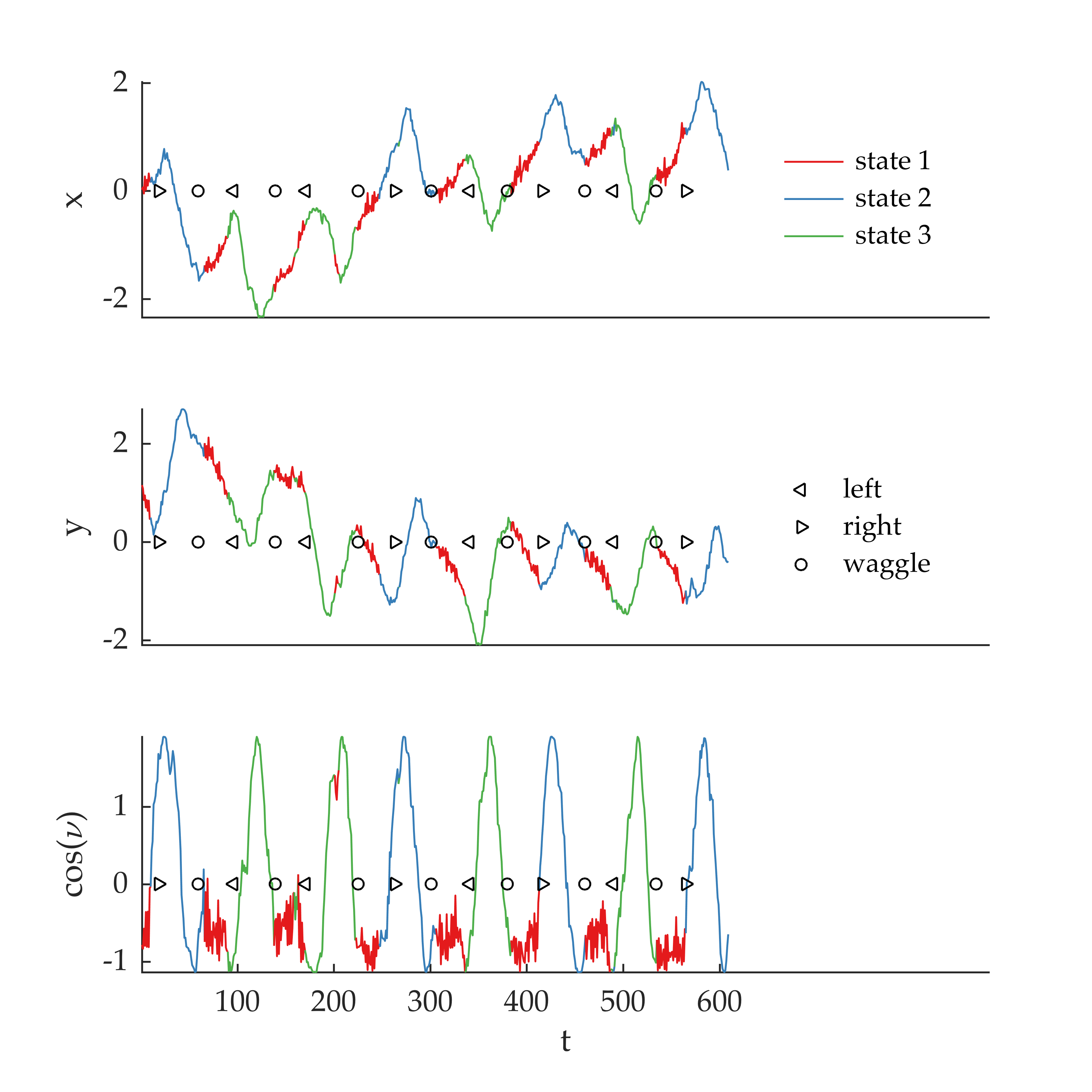}
\caption{
\textbf{Left}: ECDF of aggregated posterior samples [\textrm{agg.~post.}]
($4\times n$ samples) for $S=1$
and $S=3$, and CDF of prior $\normal(0,1)$ at the latent (top) and observation
node (bottom). \textbf{Right}: segmentation of honey bee sequence 6 as
observed in $s^\ast_1,\ldots,s^\ast_n$.  Black markers indicate true change
points. p-values correspond to KS test.}
\label{fig:bee}
\end{figure*}

\subsection{Honey bee data} \label{sec:bee}
The honey bee data consists of measurements of $(x,y)$ coordinate and
head angle ($\nu$) of 6 honey bees. The measurements are
usually translated into a 4-dimensional multivariate time series
 $(x,y,\cos(\nu),\sin(\nu))$, and modelled using a switching
linear dynamical system to capture three distinct dynamical regimes, namely,
left turn, right turn and waggle \citep{Oh2008}.
We follow this strategy, and model each time series by a switching
linear dynamical system (SLDS) \citep{Fox09} as follows:
\begin{subequations}\label{eq:slds}
\begin{align}
s_1 &= 1,\, \bfz_1\sim \normal(\bzero,\bI)\\
s_t &\sim \categorial(\bpi^{(s_{t-1})})
&\forall t=2,\ldots,n\\
\bfz_t &\sim \bA^{(s_t)}\bfz_{t-1} + \bfepsilon_t,\,
\bfepsilon_t \sim \normal(\bzero, \inv{\bQ^{(s_t)}})
&\forall t=2,\ldots,n\\
\bfx_t &\sim B\bfz_t + \bfpsi_t,\,
\bfpsi_t \sim \normal(\bzero, \inv{\bR})
&\forall t=1,\ldots,n
\end{align}
\end{subequations}
where $s_t$ can be in one of $\{1,\ldots,S\}$ states.  We assume that
$\bQ^{(\cdot)}$ (for each state) and $\bR$ are diagonal matrices with
$\gammarand(\alpha,\beta)$ prior over nonzero entries, entries of $\bA^{(\cdot)}$
(for each state) and $\bB$ originate from Gaussian distribution, and 
$\bpi^{(\cdot)}$ (for each state) follow a Dirichlet distribution, i.e.,
\begin{subequations}\label{eq:slds_prior}
\begin{align}
a^{(\cdot)}_{\cdot\cdot} &\sim \normal(0,\inv{\tau_A}),\, 
b_{\cdot\cdot} \sim \normal(0,\inv{\tau_B}) \\
\tau_A &\sim \gammarand(\alpha,\beta),\,
\tau_B \sim \gammarand(\alpha,\beta) \\
\bpi^{(\cdot)} &\sim \Dirichlet(\mathbf{1}).
\end{align}
\end{subequations}
We group $\bfs = \{s_i\}_{i=1}^n$ and $Z=\{\bfz_i\}_{i=1}^n$.
We set $\alpha = \beta = 0.001$.

We fit two models with $S=1$ (standard linear dynamical system), and $S=3$,
both with a 4 dimensional latent space. We generate a posterior sample
$(\bfs^\ast,\bZ^\ast,\bA^{(1)\ast},\ldots,\bA^{(S)\ast},
\bQ^{(1)\ast},\ldots,\bQ^{(S)\ast},\bB^\ast,\bR^\ast)$, and aggregate the
\emph{standardized} latent residuals 
\begin{align}\label{eq:bee_agg1}
\tilde{\bfz}_t=(\bQ^{(s^\ast_{t})\ast})^{0.5}(\bfz^\ast_t-\bA^{(s^\ast_t)\ast}\bfz^\ast_{t-1})\,\forall\,t=2,\ldots,n,
\end{align}
and observation residuals (or innovations)
\begin{align}\label{eq:bee_agg2}
\tilde{\bfx}_t =
(\bR^{\ast})^{0.5}(\bfx^\obs_t-\bB^{\ast}\bfz^\ast_t)\,\forall \,t=2,\ldots,n.
\end{align}

For an LDS the standard approach to model criticism is to check that the
innovations sequence is zero-mean and white (see, e.g.\
\cite[\S5.1]{candy-86}), although this is usually carried out for known or
point-estimates of the parameters, not in a Bayesian setting.  We use this
check (extended to the SLDS case) below, but also consider the latent
residuals.

First, we focus on marginal structures $\tilde{z}_{kt}$ and
$\tilde{x}_{jt}$ by pooling $k=1,2,3,4$ and $j=1,2,3,4$ together, rather than
the 4-dimensional vectors themselves as shown in Figure~\ref{fig:bee} (left).
We expect that the APSs would deviate from normality more (lower p-value) when
$S=1$, compared to $S=3$, and we observe this to be true for all honey bee
sequences except~2. For sequences 4--6, the latent segmentations of
the SLDS in terms of
$(s^\ast_1,\ldots,s^\ast_n)$ agree with the ground truth well; we present the
6th sequence in Figure~\ref{fig:bee} (right).  For sequences 1--3, we observe
that the segmentations are rather poor, similar to the results in \cite[\S
5]{Fox09}.

\begin{figure*}[t] \centering
\includegraphics[width=\textwidth]{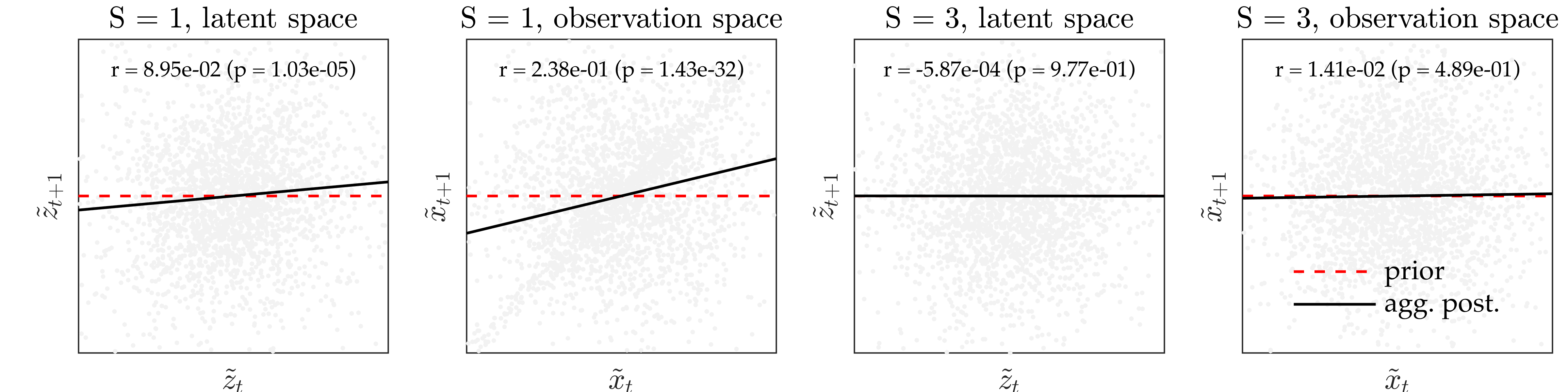} 
\caption{Scatterplot of bivariate aggregated posterior samples (607 $\times$ 4
samples) for sequence 6. The black line shows the best linear fit, whereas the
dotted line shows the expected fit in the absence of (serial) correlation. The
values in each plot are the correlation (r) and respective p-value (p).}
\label{fig:bee2} \end{figure*}

Next, we focus on the joint structures in the temporal domain by pooling,
$(\tilde{x}_{j_1 t}, \tilde{x}_{j_2 t})$ $\forall j_1,j_2 = 1,2,3,4$ and $j_1
\neq j_2$, $(\tilde{z}_{k t}, \tilde{z}_{k (t+1)})$ $\forall k = 1,2,3,4$, and
$(\tilde{x}_{j t}, \tilde{x}_{j (t+1)})$ $\forall j = 1,2,3,4$.  We expect that
this APSs would deviate from the reference distribution $\normal(\bzero,\bI_2)$ more
for $S=1$ than for $S=3$. We compute the correlation coefficients, and observe
the respective p-values for $S=1$ and $S=3$~\citep{Rahman68}. We observe that
for $S=1$, the models are rejected either in the latent domain or in the
observation domain except for sequence 3, while for $S=3$, the models are
rejected either in the latent domain or the observation domain for sequences 2
and 3 only.  In other words, for sequences 1, 4, 5 and 6, the model improves
for $S=3$, whereas for sequence 2 it fails to improve, and for sequence 3 it
degrades for $S=3$.  These observations can again be attributed to the poor
segmentation for sequences 1-3.  We show the corresponding aggregated
posteriors in the latent and observation space for sequence 6 in
Figure~\ref{fig:bee2}. We observe that the residuals in both latent and
observation space display correlations for $S=1$ while these is reduced
considerably for $S=3$. 

\begin{figure*}[t!]
\centering
\begin{minipage}[l]{0.32\textwidth}
\centering\footnotesize{(a) SE}\\
\includegraphics{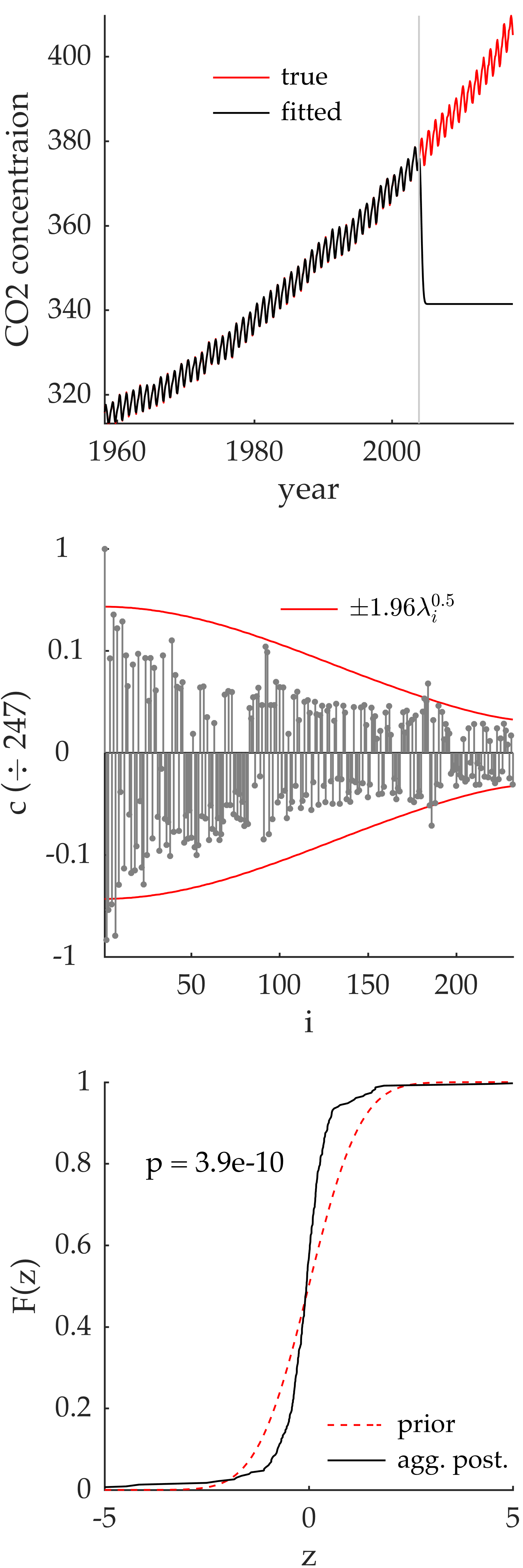}
\end{minipage}
\begin{minipage}[l]{0.32\textwidth}
\centering\footnotesize{(b) Periodic}\\
\includegraphics{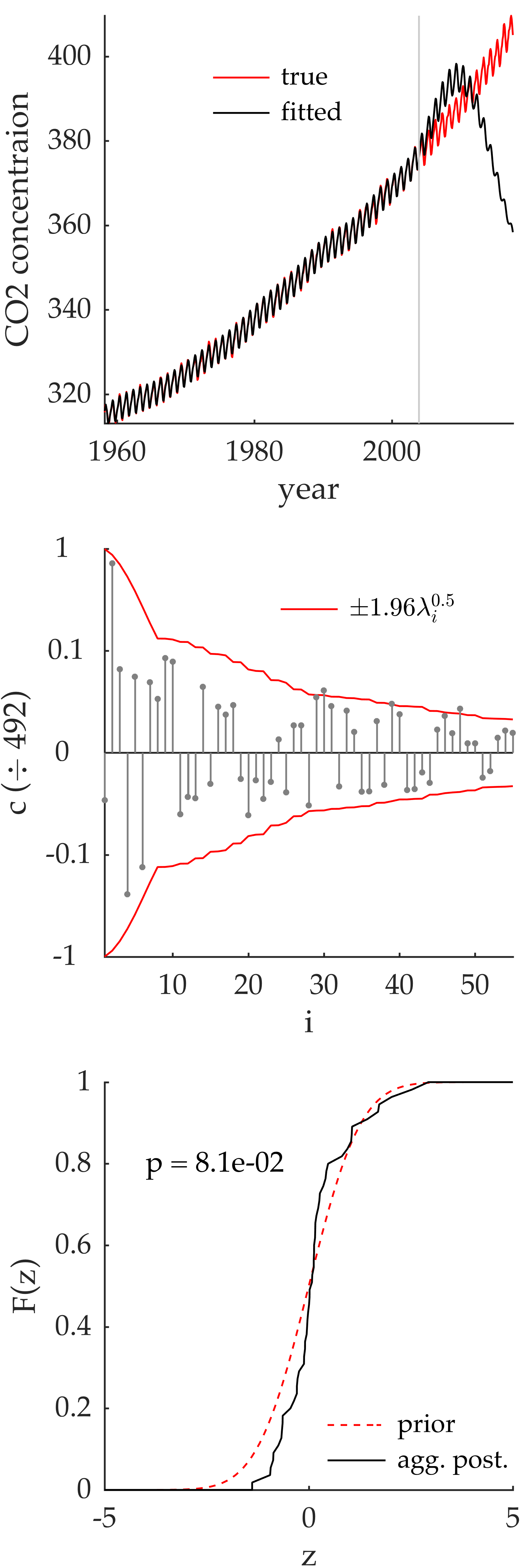}
\end{minipage}
\begin{minipage}[l]{0.32\textwidth}
\centering\footnotesize{(c) Peri.\ + SE($s$) + SE($l$)}\\
\includegraphics{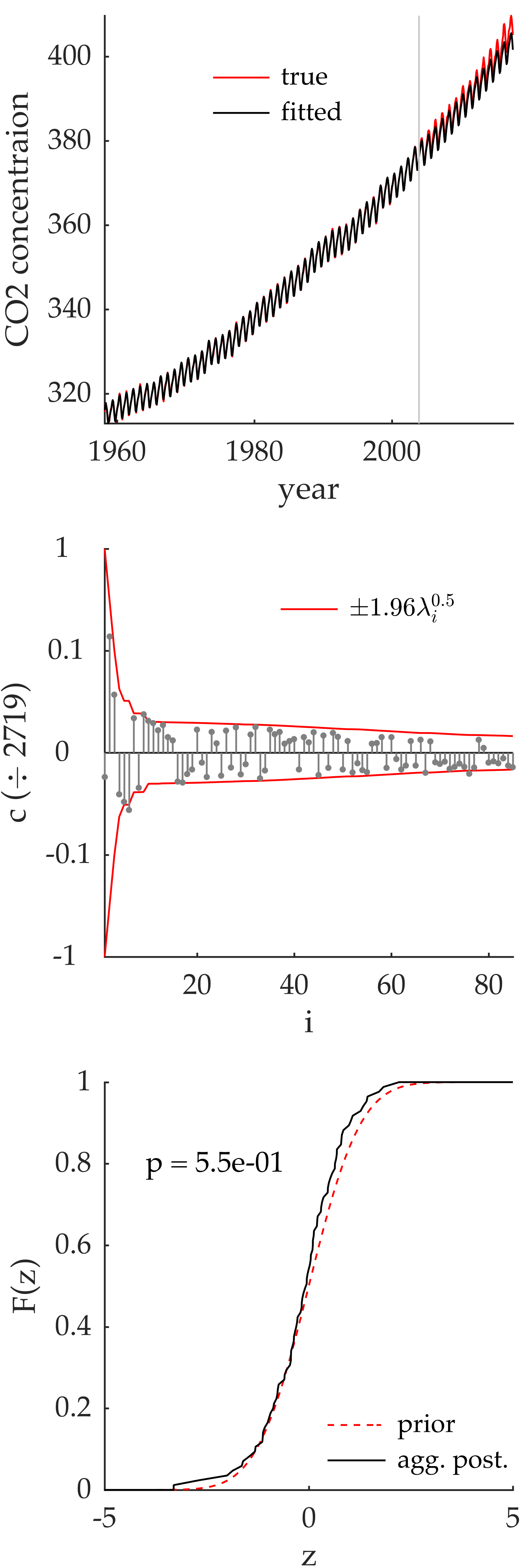}
\end{minipage}
\caption{Latent values of the Gaussian process model for CO2 emission
dataset. \textbf{Top}: Original and fitted signal. Training and testing
sets are separated by a gray line.
\textbf{Middle}: Unnormalized projections $\bfc^\ast_i$'s for $i=1,\ldots,n$,
and the respective $95\%$ confidence interval $\pm 1.96\lambda_i^{1/2}$.
We only show values for which $\lambda^\ast_i > 2 \inv{\tau^{\ast}}$. 
\emph{y-axis has been transformed by
$\mathrm{sgn}(y)|y|^{0.3}$ to show small values}.
\textbf{Bottom}: ECDFs of aggregated posterior samples [\textrm{agg.~post.}]
of the normalized projections $\bfz_i$'s and CDF of prior distribution $\mathcal{N}(0,1)$. p-values correspond to KS-test.}
\label{fig:CO2}
\end{figure*}

\subsection{Carbon emission data} \label{sec:co2}
\begin{figure*}[t!]
\begin{minipage}[r]{0.49\textwidth}
\centering\footnotesize{(a) SE}
\end{minipage}
\begin{minipage}[r]{0.49\textwidth}
\centering\footnotesize{(b) Periodic}
\end{minipage}
\includegraphics[width=\textwidth,clip=true,trim=0 0 0 0]{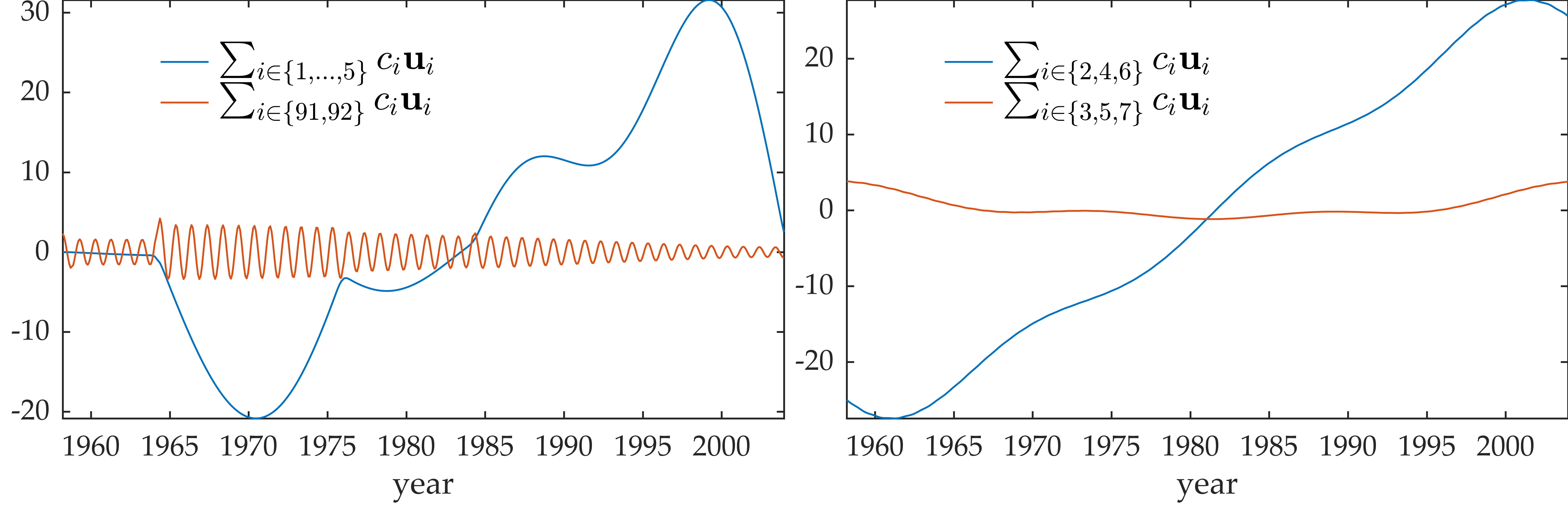}
\caption{Weighted eigenfunctions of carbon emission dataset.  The
  kinks in the plot appear due to the short length scale.}
\label{fig:CO2e}
\end{figure*}

The CO2 emission
dataset\footnote{\url{ftp://ftp.cmdl.noaa.gov/ccg/co2/trends/co2_mm_mlo.txt}}
comprises monthly average atmospheric carbon concentration $y_i$ (in parts per
million) between 1958 and 2017 (707 measurements after removing missing
values). \citet[\S 5.4.3]{Rasmussen2005} show that this time series can be
modelled well by a combination of 4 standard covariance functions involving 10
hyperparameters (and an additional parameter to model the additive white
noise).  Each covariance function is introduced to model a specific aspect of
the signal, e.g., a squared exponential kernel to model the long term trend, a
decaying periodic kernel to model the seasonal variation, a rational quadratic
kernel to model the short term irregularities, and another squared exponential
kernel to model the residual correlated noise. We show below how model
criticism and extension can be used to justify the use of covariance functions
representing similar aspects of the data.

Following Rasmussen\footnote{\url{http://learning.eng.cam.ac.uk/carl/mauna/}}
we use the measurements up to year 2004 (543) as training data and the rest
(164) as testing data\footnote{We do not use testing data for model criticism
but to show the goodness of fit visually.}. We remove the mean of the training
data before modelling, and use a zero mean function, i.e., $m(x) = 0$ or $\bfm
= \bzero$. We use
$\gammarand(\alpha,\beta)$ (See Table~\ref{table:dist})
priors over $\zeta$ and $\inv{\tau}$ to keep them positive.  We use
the GPstuff toolbox \citep{gpstuff} to generate MCMC samples, and initialize
the sampler at the maximum likelihood (ML) solution obtained using GPML
toolbox\footnote{\url{http://www.gaussianprocess.org/gpml/code/matlab/doc/}}.
We set the parameters $\alpha$ and $\beta$ such that the mean of the prior
distribution is at the ML solution, and the variance is equal to the mean.  
We generate a posterior sample $(\vartheta^\ast,\zeta^\ast,\tau^\ast\}$ and
aggregate the standardized projections 
\begin{align} \label{eq:standproj}
\bfz^\ast = \invhalf{\bLambda^{\ast}}\bfU^{\ast\top}(\bfy - \bfm^\ast)
\end{align} where $\bfU^\ast$ and $\bLambda^\ast$
are the eigenvectors and eigenvalues of kernel matrix $\bfK^\ast$ such that
$\bfK^\ast_{ij} = \kappa(x_i,x_j\g \zeta^\ast) + \inv{\tau^{\ast}}\delta(x_i,x_j)$, and
$\bfm^\ast = 0$ by design.

We first model the time series with the squared exponential or Gaussian kernel,
\begin{align} \label{eq:sekernel} \kappa_\mathrm{se}(x,x'\g \zeta) = \sigma^2_f
\exp\left(-\frac{(x-x')^2}{2l^2}\right) \end{align} where $l$ is the length
scale and $\sigma^2_f$ is the signal variance, i.e., $\zeta =
\{\sigma^2_f,l\}$. We obtain $\zeta_\textrm{ML} = (188,0.30)$ and $\zeta^\ast =
(197,0.29)$.  We present the fitted data along with unstandardized and
standardized projections in Figure~\ref{fig:CO2}\footnote{It is also possible
to model this time series with a large length scale, i.e., $\zeta = (1958,31)$
but this has lower marginal likelihood $\exp(-1198)$ as opposed to $\exp(-753)$.}. The
figure shows that the Gaussian kernel fails to model the time series as the
prediction quickly falls to the mean of the training signal and KS-test p-value
$= 4\times10^{-10}$. We observe that most of the signal strength ($\bfc_i$'s)
is concentrated at lower frequencies (corresponding to large eigenvalues
$\lambda_{i\in \{1,5\}}$). The respective eigenvectors correspond to an upward
trend. Also, a relatively high strength is observed at eigenvalues 92--93. The
respective eigenvectors correspond to sinusoids of frequency ${\sim}1$ year
(see Figure~\ref{fig:CO2e}a) which indicates a potential need of a periodic
covariance function to model this data.

To tackle this, we use the decaying periodic function to model the time series,
\citep[\S 5.4.3]{Rasmussen2005} 
\begin{align} \label{eq:pekernel}
\kappa_\mathrm{pe}(x,x'\g \zeta) = \sigma^2_f
\exp\left(-\frac{2\sin^2(\pi(x-x')/p)}{l_p^2}\right)
\exp\left(-\frac{(x-x')^2}{2l_d^2}\right)
\end{align} where $p$ is the period of the
covariance function.  Therefore, $\zeta = (\sigma^2_f,p,l_p,l_d)$. We obtain
$\zeta_\textrm{ML} = (283, 1, 5.13, 5.86)$, and $\zeta^\ast = (385, 1, 4.88,
6.09)$. We observe that this provides a better fit than squared exponential
kernel (p-value $0.08$).  Although the KS-test fails to reject the fitted
model (perhaps due to lack of samples), we observe that the signal strengths
($\bfc_i$'s) still deviate from their expected values.  In particular, the
second, fourth and sixth projections show relatively high values compared to
third, fifth and seventh. The signal $\sum_{i \in \{2,4,6\}} c_i
\bu_i$ corresponds to an upward trend, which corroborates the need to model
the trend further. See Figure~\ref{fig:CO2e}b.
Note that although the CO2 data is a time-series, 
the analysis of the $\bfc$-samples (see Eq.~\ref{eq:gp_unnorm_proj})
does not depend on this, and can also be
used where the input-space is multi-dimensional.

To accommodate the upward trend, we introduce a squared exponential kernel with
a relatively large length scale. However, to avoid modelling small scale
variations with the same kernel, we use combination of two squared exponential
kernels with two different length scales. Therefore, $\zeta =
(\sigma^2_f,p,l_p,l_d,\sigma^2_{fs},l_s,\sigma^2_{fl},l_l)$ where the last four
parameters belong to the two squared exponential kernels with small ($s$) and
large ($l$) length scales. We obtain $\zeta_\textrm{ML} =
(4.37,1,1.78,74.60,0.81,0.92,4132,27.14)$, and\\ $\zeta_\ast =
(2.25,1,1.24,73.88,0.32,0.66,4095,32.25)$.  We observe that this improves the
fit even further, both in terms of the testing data (visually) and in terms of
unstandardized projections. The KS-test uses more samples, and still fails to
reject the model (p-value $0.55$).

Model criticism of Gaussian processes in the observation space has been
discussed by \cite{lloyd2015statistical}. However, their approach is
different from the standard posterior predictive check since the authors use
hold-out data rather than using the observed data twice. Although this approach
shows if the response on hold-out data is different for the fitted model, it
does not necessarily point out how the model can be extended. 

One could generate a replicate sample from $\bfy^\rep \sim p( \cdot
  \g \zeta^\ast, \bfX^\obs)$, and compare $\bfy^\rep$ and $\bfy^\obs$
as for a posterior predictive check. However, note that in this case
$\bfy^\rep$ will be an \emph{independent} draw from the GP with parameters
$\zeta^\ast$ and input locations $\bfX^\obs$, hence it could look very different
from $\bfy^\obs$---this is why \citet{lloyd2015statistical} make use
of held-out data. Also, it would be difficult to come up with a 
suitable discrepancy function in this case. One could consider the $\chi^2$
discrepancy\footnote{Inspired by \citet[Eq.~(8)]{Gelman96posteriorpredictive}
}, i.e., $\bfy^\top \inv{\bfK} \bfy$. However, this quantity is fitted
when sampling the kernel parameters $\zeta$, and is also (as discussed
above) dominated by the noise for small eigenvalues of $\bfK$. Other 
discrepancy measures could be investigated,  but
exploring these alternatives is beyond the scope of this paper.

\section{Discussion} \label{sec:discussion}
Model criticism explores the discrepancies between a statistical model
$P(X,\unobs)$ and observed data $x^\obs$. This is often achieved by
generating replicate observations $X^\rep \sim P(X\g
x^\obs)$ from the fitted model, and investigating which
aspects $\disc(X,\unobs)$ of the replicated observations do not match the observed
data.  Instead here we have focused on pulling the effect of the data back into
the latent space, and investigating if the posterior sample $\unobss^\ast \sim
P(\unobs\g x^\obs)$ follows the prior distribution $P(\unobs)$, as it should do
by \prop~\ref{prop:prop} if the data were generated by the model.  This is tested by
aggregating related variables with the same prior distribution and comparing
them with the associated prior.

It should be noted that model criticism is not used to judge if a model is
right or wrong.  On the contrary, it is widely accepted that \emph{all models
are wrong but some are useful} \citep[p.\ 424]{box-draper-87}. Model criticism
aims at understanding the limitations of the model with the hope that a better
model can be found, e.g., since all models are basically simplifications of a
more complex process, model criticism inspects if the simplification is
meaningful, or if the statistical assumptions made are reasonable.
Following this principle, we have discussed four examples of model criticism
in latent space. We have shown that by analysing the distribution of the
aggregated posterior, a model can be extended so that the aggregated
posteriors follow the respective prior distributions better.

\subsection*{Acknowledgements}
We thank the anonymous referees for pointing out the work of
\cite{johnson2007} and \cite{YuanJohnson12}, and for comments
which helped to improve the paper.

%\bibliographystyle{plainnat}
%\bibliography{ref.bib}

\begin{thebibliography}{33}
\providecommand{\natexlab}[1]{#1}
\providecommand{\url}[1]{\texttt{#1}}
\expandafter\ifx\csname urlstyle\endcsname\relax
  \providecommand{\doi}[1]{doi: #1}\else
  \providecommand{\doi}{doi: \begingroup \urlstyle{rm}\Url}\fi

\bibitem[Bayarri and Berger(2000)]{Bayarri00}
M.~J. Bayarri and James~O. Berger.
\newblock {{p}-values for Composite Null Models}.
\newblock \emph{Journal of the American Statistical Association}, 95\penalty0
  (452):\penalty0 1127--1142, December 2000.

\bibitem[Belin and Rubin(1995)]{Belin95}
T.~R. Belin and D.~B. Rubin.
\newblock {The Analysis of Repeated-Measures Data on Schizophrenic Reaction
  Times using Mixture Models}.
\newblock \emph{Statistics in Medicine}, 14\penalty0 (8):\penalty0 747--768,
  1995.

\bibitem[Box and Draper(1987)]{box-draper-87}
G.~E.~P. Box and N.~R. Draper.
\newblock \emph{{Empirical Model-Building and Response Surfaces}}.
\newblock Wiley, 1987.

\bibitem[Box(1980)]{box-80}
G.~E.P Box.
\newblock {Sampling and Bayes' Inference in Scientific Modelling and
  Robustness}.
\newblock \emph{Journal of the Royal Statistical Society}, 143(4):\penalty0
  383--430, 1980.

\bibitem[Breusch and Pagan(1979)]{Breusch79}
T.~S. Breusch and A.~R. Pagan.
\newblock A simple test for heteroscedasticity and random coefficient
  variation.
\newblock \emph{Econometrica}, 47\penalty0 (5):\penalty0 1287--1294, 1979.

\bibitem[Buccigrossi and Simoncelli(1999)]{buccigrossi-simoncelli-99}
R.~P. Buccigrossi and E.~P. Simoncelli.
\newblock {Image Compression via Joint Statistical Characterization in the
  Wavelet Domain}.
\newblock \emph{IEEE Transactions on Signal Processing.}, 8(12):\penalty0
  1688--1701, 1999.

\bibitem[Candy(1986)]{candy-86}
J.~V. Candy.
\newblock \emph{{Signal Processing: The Model Based Approach}}.
\newblock McGraw-Hill, 1986.

\bibitem[Cook et~al.(2006)Cook, Gelman, and Rubin]{Cook06}
S.~R. Cook, A.~Gelman, and D.~B. Rubin.
\newblock Validation of software for bayesian models using posterior quantiles.
\newblock \emph{Journal of Computational and Graphical Statistics}, 15\penalty0
  (3):\penalty0 675--692, 2006.

\bibitem[Durbin and Watson(1950)]{Durbin50}
J.~Durbin and G.~S. Watson.
\newblock Testing for serial correlation in least squares regression: I.
\newblock \emph{Biometrika}, 37\penalty0 (3/4):\penalty0 409--428, 1950.

\bibitem[Fox et~al.(2009)Fox, Sudderth, Jordan, and Willsky]{Fox09}
E.~Fox, E.~B. Sudderth, M.~I. Jordan, and A.~S. Willsky.
\newblock {Nonparametric Bayesian Learning of Switching Linear Dynamical
  Systems}.
\newblock In \emph{Advances in Neural Information Processing Systems 21}, pages
  457--464. 2009.

\bibitem[Gelman et~al.(1996)Gelman, Meng, and
  Stern]{Gelman96posteriorpredictive}
A.~Gelman, X.~Meng, and H.~Stern.
\newblock Posterior predictive assessment of model fitness via realized
  discrepancies.
\newblock \emph{Statistica Sinica}, pages 733--807, 1996.

\bibitem[Gelman et~al.(2004)Gelman, Carlin, Stern, and
  Rubin]{gelman-carlin-stern-rubin-04}
A.~Gelman, J.~B. Carlin, H.~S. Stern, and D.~B. Rubin.
\newblock \emph{{Bayesian Data Analysis}}.
\newblock Chapman and Hall, London, 2004.
\newblock Second edition.

\bibitem[Gopalan et~al.(2015)Gopalan, Hofman, and Blei]{GopalanHB15}
P.~Gopalan, J.~M. Hofman, and D.~M. Blei.
\newblock Scalable recommendation with hierarchical poisson factorization.
\newblock In \emph{Proceedings of the Thirty-First Conference on Uncertainty in
  Artificial Intelligence, {UAI}}, pages 326--335, 2015.

\bibitem[Hinton et~al.(2006)Hinton, Osindero, and Teh]{hinton-osindero-teh-06}
G.~E. Hinton, S.~Osindero, and Y.~W. Teh.
\newblock {A Fast Learning Algorithm for Deep Belief Nets}.
\newblock \emph{Neural Computation}, 18:\penalty0 1527--1554, 2006.

\bibitem[Johnson(2007)]{johnson2007}
V.~E. Johnson.
\newblock Bayesian model assessment using pivotal quantities.
\newblock \emph{Bayesian Anal.}, 2\penalty0 (4):\penalty0 719--733, 12 2007.
\newblock \doi{10.1214/07-BA229}.

\bibitem[Lloyd and Ghahramani(2015)]{lloyd2015statistical}
J.~R Lloyd and Z.~Ghahramani.
\newblock {Statistical Model Criticism Using Kernel Two Sample Tests}.
\newblock In \emph{Advances in Neural Information Processing Systems}, 2015.

\bibitem[Martin et~al.(2001)Martin, Fowlkes, Tal, and Malik]{MartinFTM01}
D.~Martin, C.~Fowlkes, D.~Tal, and J.~Malik.
\newblock {A Database of Human Segmented Natural Images and its Application to
  Evaluating Segmentation Algorithms and Measuring Ecological Statistics}.
\newblock In \emph{Proceedings of 8th International Conference on Computer
  Vision}, volume~2, pages 416--423, 2001.

\bibitem[Meulders et~al.({1998})Meulders, Gelman, Van~Mechelen, and
  De~Boeck]{meulders-etal-98}
M.~Meulders, A.~Gelman, I.~Van~Mechelen, and P.~De~Boeck.
\newblock {Generalizing the Probability Matrix Decomposition Model: an Example
  of Bayesian Model Checking and Model Expansion}.
\newblock In {Hox, J. J. and de Leeuw, E. D.}, editor, \emph{Assumptions,
  Robustness and Estimation Methods in Multivariate Modeling}. TT-Publikaties,
  Amsterdam, {1998}.

\bibitem[Oh et~al.(2008)Oh, Rehg, Balch, and Dellaert]{Oh2008}
S.~M. Oh, J.~M. Rehg, T.~Balch, and F.~Dellaert.
\newblock {Learning and Inferring Motion Patterns using Parametric Segmental
  Switching Linear Dynamic Systems}.
\newblock \emph{International Journal of Computer Vision}, 77\penalty0
  (1):\penalty0 103--124, 2008.

\bibitem[O'Hagan(2003)]{ohagan-03}
A.~O'Hagan.
\newblock {HSSS Model Criticism}.
\newblock In P.~J. Green, N.~L. Hjort, and S.~Richardson, editors,
  \emph{{Highly Structured Stochastic Systems}}, pages 422--444. Oxford
  University Press, 2003.

\bibitem[Plummer(2003)]{Plummer03jags}
M.~Plummer.
\newblock {{JAGS}: A Program for Analysis of {B}ayesian Graphical Models Using
  {G}ibbs Sampling}.
\newblock In \emph{Proceedings of the 3rd International Workshop on Distributed
  Statistical Computing (DSC 2003)}, 2003.

\bibitem[Rahman(1968)]{Rahman68}
N.~A. Rahman.
\newblock \emph{{A Course in Theoretical Statistics}}.
\newblock Charles Griffin and Company, 1968.

\bibitem[Rasmussen and Williams(2006)]{Rasmussen2005}
C.~E. Rasmussen and C.~K.~I. Williams.
\newblock \emph{Gaussian Processes for Machine Learning}.
\newblock The MIT Press, 2006.
\newblock ISBN 026218253X.

\bibitem[Ratmann et~al.(2009)Ratmann, Andrieu, Wiuf, and
  Richardson]{Ratmann10576}
O.~Ratmann, C.~Andrieu, C.~Wiuf, and S.~Richardson.
\newblock {Model criticism based on likelihood-free inference, with an
  application to protein network evolution}.
\newblock \emph{Proceedings of the National Academy of Sciences}, 106\penalty0
  (26):\penalty0 10576--10581, 2009.
\newblock ISSN 0027-8424.
\newblock \doi{10.1073/pnas.0807882106}.

\bibitem[Rubin(1984)]{rubin-84}
D.~B. Rubin.
\newblock {Bayesianly Justifiable and Relevant Frequency Calculations for the
  Applied Statistician}.
\newblock \emph{{Annals of Statistics}}, 12:\penalty0 1151--1172, 1984.

\bibitem[Salakhutdinov and Mnih(2008)]{salakhutdinov2008b}
R.~Salakhutdinov and A.~Mnih.
\newblock Bayesian probabilistic matrix factorization using {M}arkov chain
  {M}onte {C}arlo.
\newblock In \emph{Proceedings of the International Conference on Machine
  Learning}, volume~25, 2008.

\bibitem[Tang et~al.(2012)Tang, Salakhutdinov, and Hinton]{Tang12}
Y.~Tang, R.~Salakhutdinov, and G.~E. Hinton.
\newblock {Deep Mixtures of Factor Analysers}.
\newblock In \emph{Proceedings of the 29th International Conference on Machine
  Learning}, 2012.

\bibitem[Vanhatalo et~al.(2013)Vanhatalo, Riihim\"{a}ki, Hartikainen,
  Jyl\"{a}nki, Tolvanen, and Vehtari]{gpstuff}
J.~Vanhatalo, J.~Riihim\"{a}ki, J.~Hartikainen, P.~Jyl\"{a}nki, V.~Tolvanen,
  and A.~Vehtari.
\newblock {GPstuff}: {Bayesian} modeling with {Gaussian} processes.
\newblock \emph{J. Mach. Learn. Res.}, 14\penalty0 (1):\penalty0 1175--1179,
  April 2013.
\newblock ISSN 1532-4435.

\bibitem[Wainwright and Simoncelli(2000)]{Wainwright99b}
M~J Wainwright and E~P Simoncelli.
\newblock {Scale Mixtures of {Gaussians} and the Statistics of Natural Images}.
\newblock In \emph{Advances in Neural Information Processing Systems},
  volume~12, pages 855--861, 2000.

\bibitem[White(1980)]{White80}
H.~White.
\newblock A heteroskedasticity-consistent covariance matrix estimator and a
  direct test for heteroskedasticity.
\newblock \emph{Econometrica}, 48\penalty0 (4):\penalty0 817--838, 1980.

\bibitem[Wilk and Gnanadesikan(1968)]{Wilk68}
M.~B. Wilk and R.~Gnanadesikan.
\newblock Probability plotting methods for the analysis of data.
\newblock \emph{Biometrika}, 55\penalty0 (1):\penalty0 1--17, 1968.

\bibitem[Yuan and Johnson(2012)]{YuanJohnson12}
Y.~Yuan and V.~E. Johnson.
\newblock {Goodness-of-fit diagnostics for Bayesian hierarchical models}.
\newblock \emph{{Biometrics}}, 68\penalty0 (1):\penalty0 156 -- 164, 2012.

\bibitem[Zoran and Weiss(2012)]{Zoran12}
D.~Zoran and Y.~Weiss.
\newblock {Natural Images, {G}aussian Mixtures and Dead Leaves}.
\newblock In F.~Pereira, C.~J.~C. Burges, L.~Bottou, and K.~Q. Weinberger,
  editors, \emph{Advances in Neural Information Processing Systems 25}, pages
  1736--1744. Curran Associates, Inc., 2012.

\end{thebibliography}

\end{document}